%% file: main.tex
\documentclass[11pt,a4paper]{article}
\usepackage{arabtex}
\usepackage{utf8}
\usepackage{multirow}
\usepackage[table,xcdraw]{xcolor}
\usepackage[normalem]{ulem}
\useunder{\uline}{\ul}{}
\usepackage[hyperref]{emnlp2020}
\usepackage{times}
\usepackage{adjustbox}
\usepackage{booktabs}
\usepackage{multirow}
\usepackage{latexsym}

\usepackage{url}
\usepackage{microtype}
\usepackage{xspace}
\usepackage{mathtools}
\usepackage{amssymb}
\usepackage{footnote}
\usepackage{tablefootnote}
\usepackage{graphicx}
\usepackage{color, colortbl}
\usepackage{xstring}
\usepackage{comment}
\usepackage{adjustbox}
\usepackage{float}
\restylefloat{table}
\usepackage{array,booktabs,makecell}
\usepackage{appendix}
\usepackage{array,booktabs,makecell}

\aclfinalcopy 
\usepackage{fancyhdr}
\usepackage{lastpage}
\usepackage{arydshln}
\title{Toward Micro-Dialect Identification in Diaglossic and Code-Switched Environments} 

\author{Muhammad Abdul-Mageed  ~~~Chiyu Zhang  ~~~AbdelRahim Elmadany 
~~~Lyle Ungar$^\dagger$  \\ 
  Natural Language Processing Lab,  University of British Columbia \\
  $^\dagger$Computer and Information Science, University of Pennsylvania \\
  \tt\{muhammad.mageed,a.elmadany\}@ubc.ca \tt chiyuzh@mail.ubc.ca \\$^\dagger$\tt ungar@cis.upenn.edu}
  
\begin{document}
\maketitle

\setcode{utf8}
\begin{abstract}
 Although prediction of dialects is an important language processing task, with a wide range of applications, existing work is largely limited to coarse-grained varieties. Inspired by geolocation research, we propose the novel task of \textit{Micro-Dialect Identification (MDI)} and introduce MARBERT, a new language model with striking abilities to predict a fine-grained variety (as small as that of a city) given a \textit{single}, \textit{short} message. For modeling, we offer a range of novel spatially and linguistically-motivated multi-task learning models. To showcase the utility of our models, we introduce a new, large-scale dataset of Arabic micro-varieties (low-resource) suited to our tasks. MARBERT predicts micro-dialects with 9.9\% F$_1$, $\sim76\times$ better than a majority class baseline. Our new language model also establishes new state-of-the-art on several external tasks.\footnote{Our labeled data and models will be available at: \url{https://github.com/UBC-NLP/microdialects}.}
\end{abstract}

\input{intro}

\input{data-collect}

\input{data-methods}

\input{models}

\input{marbert}

\input{model_generalization}
\input{msa_impact}

\input{discussion}
\input{comparisons}
\vspace*{-3mm}
\input{lit.tex}
\vspace*{-3mm}
\input{conc.tex}


\bibliography{anthology,emnlp2020}
\bibliographystyle{acl_natbib}

\appendix
\input{appendix.tex}
\end{document}

%% file: intro.tex
\section{Introduction}\label{sec:intro}
\vspace*{-2mm}

Sociolinguistic research has shown how language varies across geographical regions, even for areas as small as different parts of the same city~\cite{labove1964,trudgill1974linguistic}. These pioneering studies often used field work data from a handful of individuals and focused on small sets of carefully chosen features, often phonological. Inspired by this early work, researchers have used geographically tagged social media data from hundreds of thousands of users to predict user location~\cite{paul2011you,amitay2004web,han2014text,rahimi2017neural,binxuan2019large,tian2020twitter,zhong-etal-2020-interpreting} or to develop language identification tools~\cite{lui2012langid,zubiaga2016tweetlid,jurgens-etal-2017-incorporating,dunn2020mapping}. Whether it is possible at all to predict the \textit{micro-varieties}~\footnote{We use \textit{micro-variety} and \textit{micro-dialect} interchangeably.} of the same general language is a question that remains, to the best of our knowledge, unanswered. In this work, our goal is to investigate this specific question by introducing the \textit{novel} task of \textit{Micro-Dialect Identification} (MDI). Given a single sequence of characters (e.g., a single tweet), the goal of MDI is to predict the particular micro-variety (defined at the level of a city) of the \textit{community of users} to which the posting user belongs. This makes MDI different from geolocation in at least two ways: in geolocation, \textbf{(1)} a model consumes a \textit{collection of messages} (e.g., 8-85 messages in popular datasets~\cite{huang2019hierarchical} and \textbf{(2)} predicts the \textit{location} of the posting user (i.e., user-level). In MDI, a model takes as input a \textit{single message}, and predicts the \textit{micro-variety} of that message (i.e., message-level).

While user location and micro-dialect (MD) are conceptually related (e.g., with a tag such as \textit{Seattle} for the first, and \textit{Seattle English}, for the second), they arguably constitute two \textit{different} tasks. This is because location is an attribute of a \textit{person} who authored a Wikipedia page~\cite{overell2009geographic} or posted on Facebook~\cite{backstrom2010find} or Twitter~\cite{han2012geolocation}, whereas MD is a characteristic of \textit{language within a community of speakers} who, e.g., use similar words to refer to the same concepts in real world or pronounce certain sounds in the same way. To illustrate, consider a scenario where the same user is at different locations during different times. While a geolocation model is required to predict these \textit{different} locations (for that \textit{same} person), an MDI model takes as its target predicting the \textit{same} micro-variety for texts authored by the person (regardless of the user location). After all, while the language of a person can, and does, change when they move from one region to another, such a change takes time. 

Concretely, although to collect our data we use location as an initial \textit{proxy} for user MD, we do not just exploit data where $n$ number of posts (usually $n$=$10$) came from the location of interest (as is usually the case for geolocation). Rather, to the extent it is possible, we take the additional \textit{necessary} step of manually verifying that a user does live in a given region, and has not moved from a different city or country (at least recently, see Section~\ref{sec:data-acq}). We hypothesize that, if we are able to predict user MD based on such data, we will have an empirical evidence suggesting MD does exist and can be detected. As it turns out, while it is almost impossible for humans to detect MD (see Section~\ref{subsec:md-annot} for a human annotation study), our models predict varieties as small as those of cities surprisingly well (9.9\% F$_1$, $\sim76\times$ higher than a majority class baseline based on a single, short message) (Sections~\ref{sec:model_gen} and~\ref{sec:msa_impact}).

\textbf{Context.} MDI can be critical for multilingual NLP, especially for social media in global settings. In addition to potential uses to improve machine translation, web data collection and search, and pedagogical applications ~\cite{jauhiainen2019discriminating}, MDI can be core for essential real-time applications in health and well-being~\citep{paul2011you}, recommendation systems~\cite{quercia2010recommending}, event detection~\cite{sakaki2010earthquake}, and disaster response~\cite{carley2016crowd}. Further, as technology continues to play an impactful role in our lives, access to nuanced NLP tools such as MDI becomes an issue of equity~\cite{jurgens2017incorporating}. The great majority of the world's currently known 7,111 living languages,\footnote{Source: \url{https://www.ethnologue.com}.} however, are not NLP-supported. This limitation also applies to closely related languages and varieties, even those that are widely spoken. 

We focus on one such situation for the Arabic language, a large collection of similar varieties with $\sim$400 million native speakers. For Arabic, currently available NLP tools are limited to the standard variety of the language, Modern Standard Arabic (MSA), and a small set of dialects such as Egyptian, Levantine, and Iraqi. Varieties comprising dialectal Arabic (DA) differ amongst themselves and from MSA at various levels, including phonological and morphological~\cite{watson2007phonology}, lexical~\cite{salameh2018fine,mageedYouTweet2018,qwaider2018shami}, syntactic~\cite{benmamoun2011agreement}, and sociological~\cite{bassiouney2020arabic}. Most main Arabic dialects, however, remain understudied. The situation is even more acute for MDs, where very limited knowledge (if at all) currently exists. The prospect of research on Arabic MDs is thus large.

A major limitation to developing robust and equitable language technologies for Arabic language varieties has been absence of large, diverse data. A number of pioneering efforts, including shared tasks~\citep{zampieri2014report,malmasi2016discriminating,zampieri2018language}, have been invested to bridge this gap by collecting datasets. However, these works either depend on automatic geocoding of user profiles~\citep{mageedYouTweet2018}, which is not quite accurate, as we show in Section~\ref{sec:data-acq}; use a small set of dialectal seed words as a basis for the collection~\citep{zaghouani2018arap,qwaider2018shami}, which limits text diversity; or are based on translation of a small dataset of sentences rather than naturally-occurring text~\cite{salameh2018fine}, which limits the ability of resulting tools. The recent Nuanced Arabic Dialect Identification (NADI) shared task~\cite{mageed-etal-2020-nadi} aims at bridging this gap.

In this work, following~\newcite{gonccalves2014crowdsourcing,doyle2014mapping,sloan2015tweets}, among others, we use \textit{location as a surrogate for dialect} to build a very large scale Twitter dataset ($\sim$6 billion tweets), and automatically label a subset of it ($\sim$500M tweets) with coverage for all 21 Arab countries at the nuanced levels of state and city (i.e., micro-dialects). In a departure from geolocation work, we then \textit{manually} verify user locations, excluding $\sim37\%$ of users. We then exploit our data to develop highly effective hierarchical and multi-task learning models for detecting MDs.


Other motivations for choosing Arabic as the context for our work include that \textbf{(1)} Arabic is a \textit{diaglossic} language~\cite{ferguson1959diglossia,bassiouney2020arabic} with a so-called `High' variety (used in educational and formal settings) and `Low' variety (used in everyday communication). This allows us to exploit dialgossia in our models. In addition, \textbf{(2)} for historical reasons, different people in the Arab world \textit{code-switch} in different foreign languages (e.g., English in Egypt, French in Algeria, Italian in Libya). This affords investigating the impact of code-switching on our models, thereby bringing yet another novelty to our work. Further, \textbf{(3)} while recent progress in \textit{transfer learning} using language models such as BERT~\cite{devlin2018bert} has proved strikingly useful, Arabic remains dependent on multilingual models such as mBERT 
trained on the restricted Wikipedia domain with limited data. Although an Arabic-focused language model, AraBERT~\cite{baly2020arabert}, was recently introduced, it is limited to MSA rather than dialects. This makes AraBERT sub-optimal for social media processing as we show empirically. We thus present a novel Transformer-based Arabic language model, MARBERT, for MDs. Our new model exploits a massive dataset of 1B posts, and proves very powerful: It establishes new SOTA on a wide range of tasks. Given the impact self-supervised language models such as BERT have made, our work has the potential to be a key milestone in \textit{all} Arabic (and perhaps multilingual) NLP. 

To summarize, we offer the following contributions: (1) We collect a massive dataset from Arabic social media and exploit it to develop a large human-labeled corpus for Arabic MDs. (2) For modeling, we introduce a novel, \textit{spatially} motivated hierarchical attention multi-task learning (HA-MTL) network that is suited to our tasks and that proves highly successful. (3) We then introduce \textit{linguistically} guided multi-task learning models that leverage the diaglossic and code-switching environments in our social data. (4) We offer a new, powerful Transformer-based language model trained with self-supervision for Arabic MDs. (5) Using our powerful model, we establish new SOTA results on several external tasks.

The rest of the paper is organized as follows: In Section~\ref{sec:data-acq}, we introduce our methods of data collection and annotation. Section~\ref{sec:exp-data} is about our experimental datasets and methods. We present our various models in Section~\ref{sec:sup-exps} and our new micro-dialectal model, MARBERT, in Section~\ref{sec:marbert}. We investigate model generalization in Section~\ref{sec:model_gen}, and the impact of removing MSA from our data in Section~\ref{sec:msa_impact}. Section~\ref{sec:discuss} is a discussion of our findings. We compare to other works in Section~\ref{sec:comp}, review related literature in Section~\ref{sec:lit}, and conclude in Section~\ref{sec:conc}.

%% file: data-collect.tex
\section{Data Acquisition and Labeling}\label{sec:data-acq}
\vspace*{-2mm}
We first acquire a large user-level dataset covering the whole Arab world. We then use information in user profiles (available only for a subset of users) to automatically assign city, state, and country labels to each user. Since automatic labels can be noisy (e.g., due to typos in city names, use of different languages in user profiles), we manually fix resulting errors. To further account for issues with human mobility (e.g., a user from one country moving to another), we manually inspect user profiles, tweets, and network behavior and verify assigned locations. Finally, we propagate city, state, and country labels from the user to the tweet level (each tweet gets the label assigned to its user). We now describe our data methods in detail. 

\subsection{A Large User-Level, Tagged Collection}
\vspace*{-3mm}
\begin{figure}[h]
\center 
\frame{\includegraphics[width=\columnwidth,height=4.5cm]{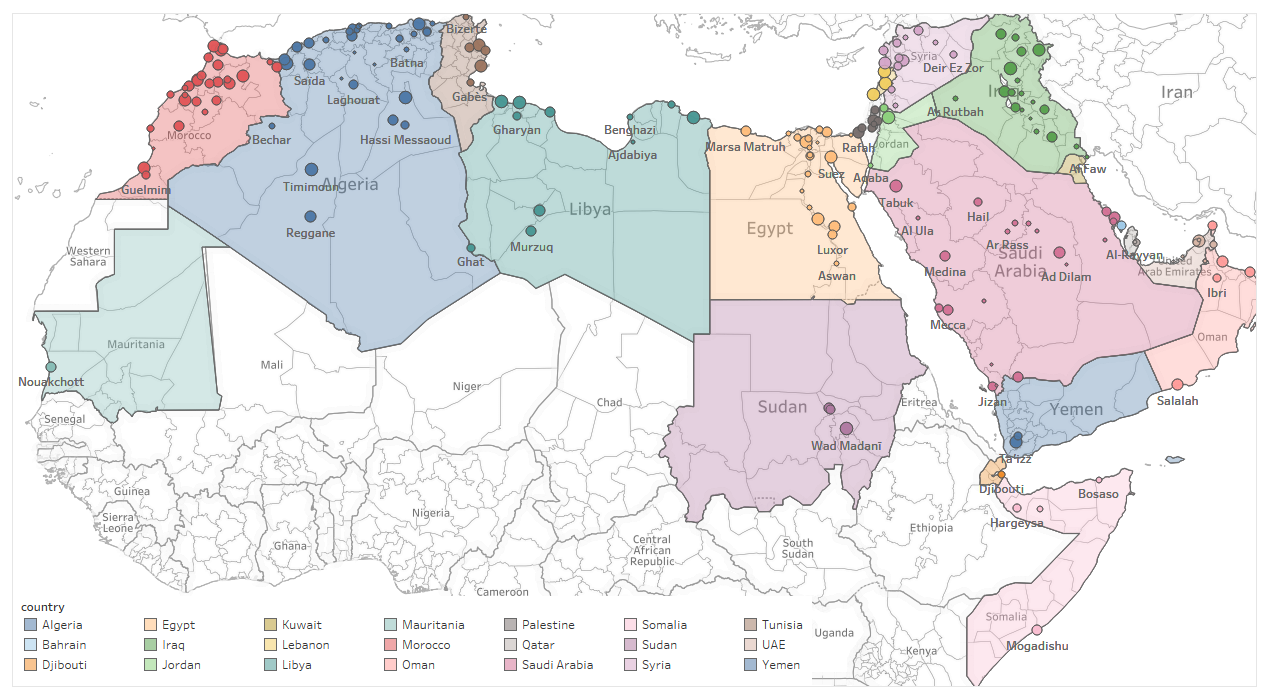}}
\caption{All 21 Arab countries in our data, with states demarcated in thin black lines within each country. All 319 cities from our user location validation study, in colored circles, are overlayed within respective states.} \vspace*{-2mm}
\label{fig:all_loc}
\end{figure}

To develop a large scale dataset of Arabic varieties, we use the Twitter API to crawl up to 3,200 tweets from $\sim$2.7 million users collected from Twitter with bounding boxes around the Arab world. Overall, we acquire $\sim$6 billion tweets.
We then use the Python geocoding library \textit{geopy} to identify user location in terms of countries (e.g., Morocco) and cities (e.g., Beirut).\footnote{Geopy (\url{https://github.com/geopy}) is a client for several popular geocoding web services aiming at locating the coordinates of addresses, cities, countries, and landmarks across the world using third-party geocoders. In particular, we use the Nominatim geocoder for OpenStreetMap data (\url{https://nominatim.openstreetmap.org}). With Nominatim, geopy depends on user-provided geographic information in Twitter profiles such as names of countries or cities to assign user location.} Out of the 2.7 million users, we acquired \textit{both} `city' and `country' label for $\sim$233K users who contribute $\sim$ 507M tweets. The total number of cities initially tagged was 705, but we manually map them to only 646 after correcting several mistakes in results returned by geopy. Geopy also returned a total of 235 states/provinces that correspond to the 646 cities, which we also manually verified. We found all state names to be correct and to correspond to their respective cities and countries.\footnote{More information about manual correction of city tags is in Section~\ref{subsec:append:city_tag} in the Appendix.} 

\subsection{Validation of User Location}\label{subsec:usr-valid}

Even after manually correcting location labels, it cannot be guaranteed that a user \textit{actually} belongs to (i.e., is a local of) the region (city, state, and country) they were assigned. Hence, we manually verify user location through an annotation task. To the extent it is possible, this helps us avoid assigning false MD labels to users whose profile information was captured rightly in the previous step but who indeed are \textit{not} locals of automatically labeled places. Before location verification, we exclude \textit{cities} that have $< 500$  tweets and \textit{users} with $<30$ tweets from the data. This initially gives us 417 cities. We then ask two native Arabic annotators to consider the automatic label for each task (city and country)~\footnote{Note that we have already manually established the link between states and their corresponding cities and countries.} and assign one label from the set \textit{\{local, non-local, unknown\}} per task for each user in the collection. We provide annotators with links to users' Twitter profiles, and instruct them to base their decisions on each user's network and posting content and behavior. As a result, we found that 81.00\% of geopy tags for country are correct, but only 62.29\% for city. This validates the need for the manual verification. Ultimately, we could verify a total of 3,085 users for \textit{both} country and city from all 21 countries but only from 319 cities.\footnote{More information about manual user verification is in Section~\ref{subsec:append:user-loc} of the Appendix.} Figure~\ref{fig:all_loc} shows a map of all 21 Arab countries, each divided into its states with cities overlayed as small colored circles. 

\subsection{Can Humans Detect Micro-Dialect?}\label{subsec:md-annot}

We were curious to know whether humans can identify micro-dialect from a \textit{single message}, and so we performed a small annotation study. We extracted a random set of 1,050 tweets from our labeled collection and asked two native speakers from two non-neighboring Arab countries to tag each tweet with a \textit{country} then (choosing from a drop-down menu) a \textit{state} label. Annotators found the state-level task very challenging (or rather ``impossible", to quote one annotator) and so we did not complete it. Hence, we also did not go to the level of \textit{city} since it became clear it will be almost impossible for humans. For country, annotators reported trying to identify larger regions (e.g. Western Arab world countries), then pick a specific country (e.g., Morocco). To facilitate the task, we asked annotators to assign an ``unknown" tag when unsure. We calculated inter-annotator agreement and found it at Cohen's~\cite{cohen1960coefficient} Kappa \textit{(K)=}$0.16$ (``poor" agreement). When we calculate the subset of data where both annotators assigned an actual country label (i.e., rather than ``unknown"; \textit{n=}$483$ tweets), we found the Kappa \textit{(K)} to increase to $0.47$ (``moderate" agreement). Overall, the annotation study emphasizes challenges humans face when attempting to identify dialects (even at the level of country sometimes).



%% file: data-methods.tex
\vspace*{-2mm}
\section{Datasets and Methods}\label{sec:exp-data}
\vspace*{-2mm}
\subsection{Datasets}
\vspace*{-2mm}
\textbf{Preprocessing}. To keep only high-quality data, we remove all retweets, reduce all consecutive sequences of the same character to only 2, replace usernames with $<$USER$>$ and URLs with $<$URL$>$, and remove all tweets with less than 3 Arabic words. This gives $\sim 277.4$K tweets. We tokenize input text only lightly by splitting off punctuation.\footnote{For most DA varieties, there are no available tokenizers.} 
Ultimately, we extract the following datasets for our experiments:

\textbf{Micro-Ara (Monolingual).} Extracted from our manually verified users, this is our core dataset for modeling. We randomly split it into 80\% training (TRAIN), 10\% development (DEV), and 10\% test (TEST). To limit GPU time needed for training, we cap the number of tweets in our TRAIN in any given country at 100K. We describe the distribution of classes in Micro-Ara in Tables~\ref{tab:data_stats} and~\ref{tab:gold_data} in the Appendix. 
We note that Micro-Ara is reasonably balanced. Table~\ref{tab:data_splits} shows our data splits. 





\begin{figure}[h]
\center 
\frame{\includegraphics[width=6.2cm]{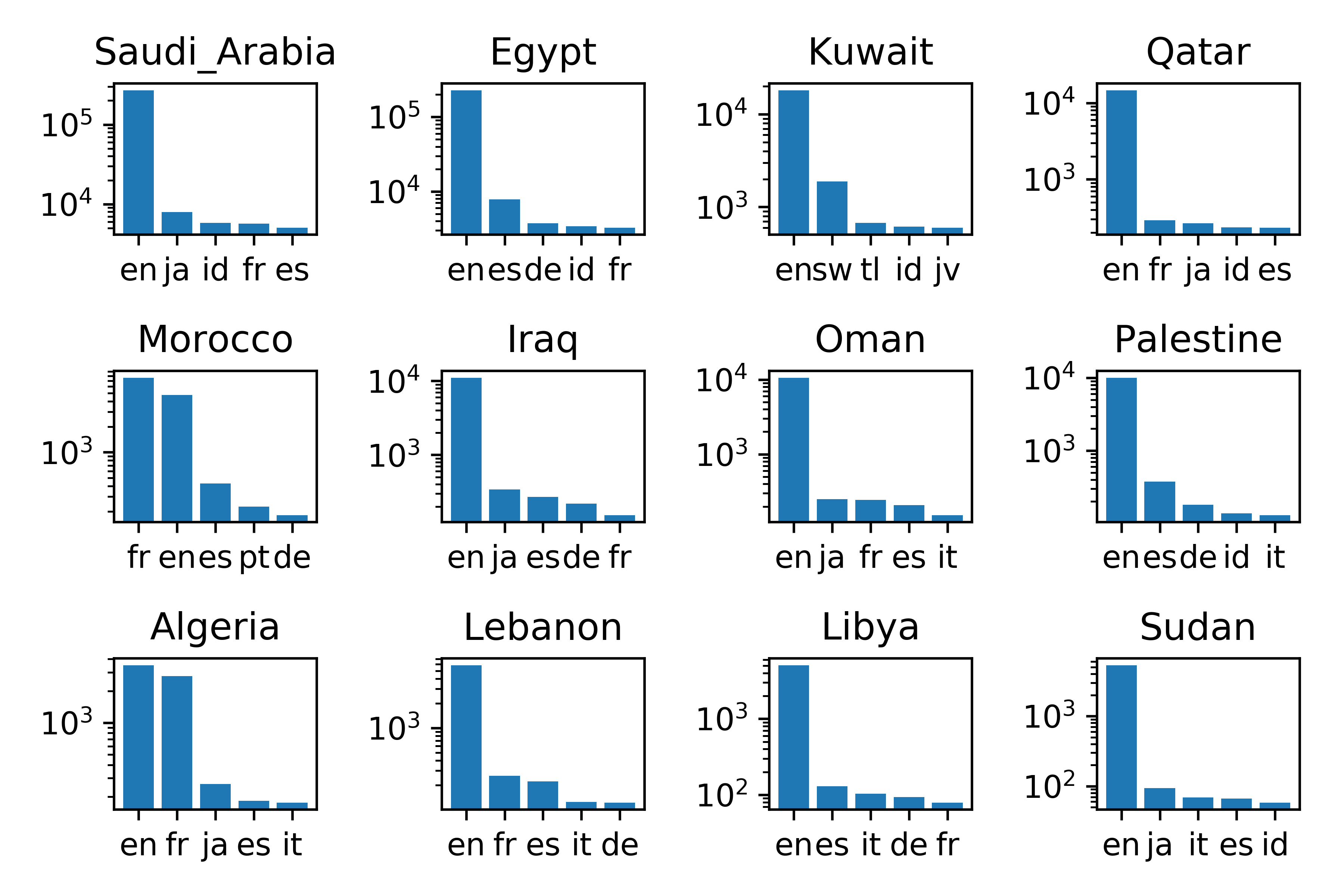}}
\caption{Code-switching over select countries (with different code-switching profiles) in CodSw.}
\vspace*{-5mm}
\label{fig:langid}
\end{figure}

\textbf{CodSw (Code-Switching).} As explained in Section~\ref{sec:intro}, Arabic speakers code-switch to various foreign languages. We hypothesize the distribution of foreign languages will vary across different regions (which proves to be true, as we show in Figure~\ref{fig:langid}), thereby providing modeling opportunities that we capture in a multi-task setting (in Section~\ref{subsec:mtbert}). Hence, we introduce a code-switching dataset (CodSw) by tagging the non-Arabic content in all tweets in our wider collection with the langid tool~\cite{lui2012langid}. Keeping only tweets with at least 3 Arabic words and at least $4$ non-Arabic words, we acquire $\sim 934$K tweets. CodSw is diverse, with a total of $87$ languages. We split CodSw as is shown in Table~\ref{tab:data_splits}. 


\textbf{DiaGloss (Diaglossia).}\label{subsec:DiaGloss_data}
We also explained in Section~\ref{sec:intro} that Arabic is a diaglossic language, with MSA as the ``High" variety and dialects as the ``Low" variety. MDs share various linguistic features (e.g., lexica) with MSA, but to varying degrees. We use an auxiliary task whose goal is to tease apart MSA from dialectal varieties. We use existence of diacritics (at least 5) as a proxy for MSA,\footnote{Unlike dialect, MSA is usually diacritized.} and direct responses (vs. tweets or retweets) as a surrogate for dialectness. In each class, we keep 500K tweets, for a total of 1M tweets split as in Table~\ref{tab:data_splits}.
We refer to this dataset as DiaGloss.

\begin{table}[h]
\footnotesize  
\centering 
\begin{tabular}{lrrr}
\hline
\multicolumn{1}{c}{\textbf{Datasets}} & \multicolumn{1}{c}{\textbf{Train}} & \multicolumn{1}{c}{\textbf{Dev}} & \multicolumn{1}{c}{\textbf{Test}} \\ \hline
Micro-Ara                     & 1,099,711      & 202,509      & 202,068       \\ 
CodSw                        & 747,173        & 93,431       & 93,565         \\ 
DiaGloss                       & 800K        & 100K      & 100K        \\
\hline
\end{tabular}
\caption{Splits of our datasets. \textbf{Micro-Ara:} City-verified dataset for MDs. \textbf{CodSw:} Code-switching dataset from our automatically-tagged collection. \textbf{DiaGloss:} MSA vs. DA data to approximate diaglossia.}\label{tab:data_splits} \vspace*{-1mm}
\end{table} \vspace*{-3mm}

\vspace*{-3mm}
\subsection{Methods} 
\textbf{BiGRUs and BERT.} We perform dialect identification at the country, state, and city levels. We use two main neural network methods, Gated Recurrent Units (GRUs)~\cite{cho2014learning}, a variation of Recurrent Neural Networks (RNNs), and Google's Bidirectional Encoder Representations from Transformers (BERT)~\cite{devlin2018bert}.  We model each task independently, but also under multi-task conditions. 

\textbf{Multi-Task Learning.}
\textit{Multi-Task Learning (MTL)} is based on the intuition that many real-world tasks involve predictions about closely related labels or outcomes. For related tasks, MTL helps achieve inductive transfer between the various tasks by leveraging additional sources of information from some of the tasks to improve performance on the target task~\cite{caruana1993}. By using training signals for related tasks, MTL allows a learner to prefer hypotheses that explain more than one task~\cite{caruana1997multitask} and also helps regularize models. In some of our models, we leverage MTL by training a \textit{single} network for our city, state, and country tasks where network layers are shared but with an independent output for each of the 3 tasks.

%% file: models.tex
\section{Models}\label{sec:sup-exps}
\vspace*{-2mm}
Here, we describe our baselines and present our MDI models. These are (i) our single- and multi-task BiGRU models (Sections~\ref{subsec:baselines},~\ref{subsec:mtl}, and~\ref{subsec:ht_mtl}), (ii) single-task BERT (Section~\ref{subsec:stbert}),  and (iii) multi-task BERT (Section ~\ref{subsec:mtbert}).

\subsection{Baselines}\label{subsec:baselines}
For all our experiments, we remove diacritics from the input text. We use two baselines: the majority class in TRAIN (Baseline I) and a single-task BiGRU (Baseline II, described below). 
For all our experiments, we tune model hyper-parameters and identify best architectures on DEV. We run all models for 15 epochs (unless otherwise indicated), with early stopping `patience' value of $5$ epochs, choosing the model with highest performance on DEV as our best model. We then run each best model on TEST, and report \textit{accuracy} and \textit{macro $F_{1}$ score}.\footnote{We include tables with results on DEV in Section~\ref{sec:append:models} in the Appendix.}

\textbf{Single-Task BiGRUs.} 
As a \textit{second baseline} (Baseline II), we build 3 independent networks (each for one of the 3 tasks) using the same architecture and model capacity. Each network has 3 hidden BiGRU layers, with 1,000 units each. More information about each of these networks and how we train them is in Section~\ref{subsec:single_task_BiGrus} in the Appendix. Table~\ref{tab:sprvsd_eval} presents our results on TEST.


\subsection{Multi-Task BiGRUs}\label{subsec:mtl}


With MTL, we design a single network to learn the 3 tasks simultaneously. In addition to our hierarchical attention MTL (HA-MTL) network, we design two architectures that differ as to how we endow the network with the \textit{attention mechanism}. We describe these next. We provide illustrations of our MTL networks in Figures~\ref{fig:mtl_t_spec_attn} and~\ref{fig:mtl2b} in the Appendix.

\textbf{Shared and Task-Specific Attention.}
We first design networks with attention at the same level in the architecture. Note that we use the same hyper-parameters as the single-task networks. We have two configurations:

\textbf{Shared Attention.}
This network has 3 hidden BiGRU layers, each of which has 1,000 units per layer (500 in each direction).\footnote{Again, 4 hidden-layered network for both the \textit{shared} and \textit{task-specific} attention settings were sub-optimal and so we do not report their results here.} All the 3 layers are shared across the 3 tasks, including the third layer. Only the third layer has attention applied. We call this setting \textit{MTL-common-attn}. 

\textbf{Task-Specific Attention.}
This network is similar to the previous one in that the first two hidden layers are shared, but differs in that the third layer (attention layer) is task-specific (i.e., independent for each task). We call this setting \textit{MTL-spec-attn}. 
This architecture will allow each task to specialize its own attention within the same network. As Table~\ref{tab:sprvsd_eval} shows, both \textit{MTL-common-attn} and \textit{MTL-spec-attn} improve over each of the two baselines (with first performing generally better).

\subsection{Hierarchical Attention MTL (HA-MTL)}\label{subsec:ht_mtl}

Instead of a `flat' attention, we turn to hierarchical attention (spatially motivated, e.g., by how a smaller regions is a part of a larger one): We design a single network for the 3 tasks but with supervision at different layers. Overall, the network has 4 BiGRU layers (each with a total of 1,000 units), the bottom-most of which has no attention. Layers 2, 3, and 4 each has dot-product attention applied, followed directly by one task-specific fully-connected layer with softmax for class prediction. 
In the two scenarios, state is supervised at the middle layer. These two architectures allow information flow with different granularity: While the city-first network tries to capture what is in the physical world a more fine-grained level (city), the country-first network does the opposite. Again, we use the same hyper-parameters as the single-task and MTL networks, but we use a dropout rate of 0.70 since we find it to work better. As Table~\ref{tab:sprvsd_eval} shows, our proposed HA-MTL models significantly outperform single-task and other BiGRU MTL models. They also outperform our Baseline II with $12.36\%$, $10.01\%$, and $13.22\%$ acc on city, state, and country prediction respectively, thus demonstrating their effectiveness on the task.

\begin{table}[h]
\centering
\begin{adjustbox}{width=\columnwidth}
\begin{tabular}{l|rr|rr|rr}
\hline
\multicolumn{1}{c|}{\textbf{Setting}} & \multicolumn{2}{c|}{\textbf{City}} & \multicolumn{2}{c|}{\textbf{State}} & \multicolumn{2}{c}{\textbf{Country}} \\ \hline
\multicolumn{1}{c|}{\textbf{Eval Metric}} & \textbf{acc} & \textbf{F1} & \textbf{acc} & \textbf{F1} & \textbf{acc} & \textbf{F1} \\ \hline
Baseline I & 1.31 & 0.01 & 3.11 & 0.03 & 9.19 & 0.80 \\
Baseline II  & 1.65 & 0.25 & 6.13 & 1.92 & 31.14 & 15.84 \\ \hdashline
MTL (common-attn) & 2.86 & 0.74 & 5.12 & 1.01 & 26.51 & 12.41 \\
MTL (spec-attn) & 2.40 & 0.68 & 4.60 & 0.90 & 27.04 & 10.98 \\
HA-MTL (city $1^{st}$) & \textbf{14.01} & \textbf{14.02} & \textbf{16.14} & \textbf{15.90} & \textbf{44.36} & 32.14 \\
HA-MTL (cntry $1^{st}$) & 13.23 &	13.06 & 15.84	& 15.40 & 44.17	& \textbf{32.37} \\\hdashline
mBERT & 19.33 & 19.45 & \textbf{21.24} & 21.67 & 47.74 & 38.12\\
MTL-mBERT (DiGls)& \textbf{19.88} & \textbf{20.11} & 21.04 & \textbf{21.69} & 48.30 & \textbf{38.34}\\
MTL-mBERT (CodSw)& 19.47 & 19.86 & 20.76 & 21.47 & \textbf{48.61} & 38.20 \\ 
\hline 
\end{tabular}%
\end{adjustbox}
\caption{Performance on TEST. \textbf{Baseline I:} majority in TRAIN. \textbf{Baseline II:} single task Attn-BiGRU. \textbf{MTL-mBERT (DiGls):} Multi-Task BERT with diaglossia. \textbf{CodSw:} code-switching.}
\label{tab:sprvsd_eval}
\end{table} \vspace*{-3mm}

\subsection{Single-Task BERT}\label{subsec:stbert}
We use the BERT-Base, Multilingual Cased model released by the authors.\footnote{\url{https://github.com/google-research/bert/blob/master/multilingual.md}.} 
For fine-tuning, we use a maximum sequence length of 50 words and a batch size of $32$. We set the learning rate to 2e-5. We train for $15$ epochs, as mentioned earlier. As Table~\ref{tab:sprvsd_eval} shows, BERT performs consistently better on the three tasks. It outperforms the best of our two HA-MTL networks with an acc of $5.32\%$ (city), $5.10\%$ (state), $3.38\%$ (country). 
To show how a small network (and hence deployable on machines with limited capacity with quick inference time) can be trained on knowledge acquired by a bigger one, we distill~\cite{hinton2015distilling,tang2019distilling} BERT representation (big) with a BiGRU (small). We provide related results in Section ~\ref{subsec:distilbert} in the Appendix.

\subsection{Multi-Task BERT}\label{subsec:mtbert}
We investigate two linguistically-motivated auxiliary tasks trained with BERT, as follows:

\textbf{Exploiting Diaglossia.} As Table~\ref{tab:sprvsd_eval} shows, a diaglossia-based auxiliary task improves over the single task BERT for both city ($0.55\%$ acc) and country ($0.56\%$ acc).

\textbf{Exploiting Code-Switching.}
We run 4 experiments with our CodSw dataset, as follows: \textbf{(1)} with the two tasks supervised at the city level, \textbf{(2)} at the country level, \textbf{(3 \& 4)} with the levels reversed (city-country vs. country-city). Although the code-switching dataset is automatically labeled, we find that when we supervise with its country-level labels, it helps improve our MDI on city ($0.14\%$ acc) and on country ($0.87\%$ acc). Related results are shown in Table~\ref{tab:sprvsd_eval}. We now describe our new language model, MARBERT. 


%% file: marbert.tex
\section{MARBERT: A New Language Model}\label{sec:marbert}
\vspace*{-2mm}

We introduce MARBERT, a new language model trained with self-supervision on 1B tweets from from our unlabaled Twitter collection (described in Section~\ref{sec:data-acq}). We train MARBERT on 100K wordPiece vocabulary, for 14 epochs with a batch size of 256 and a maximum sequence length of 128. Training took 14 days on 8 Google Cloud TPUs. We use the same network architecture as mBERT, but without the next sentence prediction objective since tweets are short. 
MARBERT has much larger token count than BERT~\cite{devlin2018bert} (15.6B vs. 3.3B), and is trained on $5\times$  bigger data than AraBERT ~\cite{baly2020arabert} (126GB vs. 24GB of text). Unlike AraBERT, which is focused on MSA, MARBERT has diverse coverage of dialects in addition to MSA. 
As Table~\ref{tab:ourLMs} shows, MARBERT significantly outperforms \textit{all} other models across the 3 tasks, with improvements of $4.13\%$ and $3.54\%$ acc over mBERT and AraBERT respectively on the country task. 
We also run a set of \textbf{MTL experiments} with MARBERT, fine-tuning it with a diaglossia auxiliary task, a code-switching auxiliary task, and \textit{both} diaglossia and code-switching as auxiliary tasks. As Table~\ref{tab:ourLMs} shows, MTL does not bring acc improvements, and helps the country task only slightly (0.30\% acc gain with CodSW). This reflects MARBERT's already-powerful representation, with little need for MTL. We provide an \textbf{\textit{error analysis}} of MARBERT's MDI in Section~\ref{sec:append:error_analysis} of the Appendix. 

\begin{table}[h]
\centering
\begin{adjustbox}{width=\columnwidth}
\begin{tabular}{l|ll|ll|ll}
\hline
\multicolumn{1}{c|}{\textbf{Setting}} & \multicolumn{2}{c|}{\textbf{City}} & \multicolumn{2}{c|}{\textbf{State}} & \multicolumn{2}{c}{\textbf{Country}} \\ \hline
\multicolumn{1}{c|}{\textbf{Eval Metric}} & \textbf{acc} & \textbf{F1} & \textbf{acc} & \textbf{F1} & \textbf{acc} & \textbf{F1} \\ \hline
mBERT (\newcite{devlin2018bert}) & 19.33 & 19.45 & 21.24 & 21.67 & 47.74 & 38.12\\ 
AraBERT (\newcite{baly2020arabert}) &18.82 &	18.73 & 20.73 &20.87 & 48.33  & 38.09
\\ \hline
MARBERT (Ours) & \textbf{20.78} & \textbf{20.41} &\textbf{22.97} & \textbf{22.58} & \textbf{51.87} & \textbf{42.17}\\
\hdashline
MTL-MARBERT (DiaGloss)& 20.19 & \textbf{20.60} & \textbf{23.22} & 22.97 & 51.53 & 41.75\\
MTL-MARBERT (CodSw) & \textbf{20.77}  & 20.56  & 23.21 & \textbf{23.16} & 51.78 & \textbf{42.36} \\ 
MTL-MARBERT (CSD) & 20.76 & 20.23 & 23.18 & 23.15 & \textbf{52.17}& 42.27 \\ 
\hline 

\end{tabular}%
\end{adjustbox}
\caption{MARBERT (ours) outperforms mBERT and AraBERT (TEST results). \textbf{CSD:} DiaGloss+CodSw.}
\label{tab:ourLMs}
\end{table}

%% file: model_generalization.tex
\vspace*{-7mm}
\section{Model Generalization}\label{sec:model_gen}
\vspace*{-2mm}
For the experiments reported thus far, we have split our Micro-Ara (monolingual) dataset randomly at the tweet level. While this helped us cover the full list of our 319 cities, including cities from which we have as few as a single user, this split does not prevent tweets from the \textit{same} user to be divided across TRAIN, DEV, and TEST. In other words, while the tweets across the splits are unique (not shared), users who posted them are not unique. We hypothesize this may have the consequence of allowing our models to acquire knowledge about user identity (idiolect) that interact with our classification tasks. To test this hypothesis, we run a set of experiments with different data splits where users in TEST are \textit{never} seen in TRAIN. To allow the model to see enough users during training, we split the data only into TRAIN and TEST and use no DEV set. We use the same hyper-parameters identified on previous experiments. An exception is the number of epochs, where we report the best epoch identified on TEST. To alleviate the concern about absence of a DEV set, we run each experiment 3 times. Each time we choose a different TEST set, and we average out performance on the 3 TEST sets. This is generally similar to cross-validation.

For this set of experiments, we first remove all cities from which we have only one user (79 cities) and run experiments across 3 different settings (\textbf{narrow}, \textbf{medium}, and \textbf{wide}).\footnote{We give each setting a name that reflects its respective geographical coverage. For example, \textit{wide} has wider coverage than \textit{medium}, which in turn has wider coverage than \textit{narrow}.} We provide a description of these 3 settings in  Section~\ref{subsec:narrow_med_wide} in the Appendix. For the \textit{narrow} setting only, we also run with the same code-switching and diaglossic auxiliary tasks (individually and combined) as before. We use mBERT fine-tuned on each respective TRAIN as our baseline for the current experiments.  

As Table~\ref{tab:model_generalization} shows, MARBERT significantly ($p<0.01$) outperforms the strong mBERT baseline across the 3 settings. With the \textit{narrow} setting on MDs, MARBERT reaches $8.12\%$ F$_1$ ($61$ cities). These results drop to $5.81\%$ (for \textit{medium}, $116$ cities) and $3.59\%$ (for \textit{wide}, $240$ cities).  
We also observe a positive impact\footnote{But not significant ($p<0.07$ for city-level with \textit{narrow}), since MARBERT is already powerful.} from the combined code-switching and diaglossic auxiliary tasks. All results are also \textit{several folds better} than a majority class city baseline (city of Abu Dhabi, UAE; not shown in Table~\ref{tab:model_generalization}). For example, results acquired with the two (combined) auxiliary tasks are $4.7\times$ better in acc and $229\times$ better for F${_1}$ than the majority class.

Importantly, although results in Table~\ref{tab:model_generalization} are not comparable to those described in Section~\ref{sec:marbert} (due to the different data splits), these results suggest that our powerful transformer models in Section~\ref{sec:marbert} may have made use of user-level information (which may have caused  inflated performance). To further investigate this issue in a reasonably comparable set up, we apply the models based on the narrow, medium and wide settings \textit{and} our single task MARBERT model (shown in Table~\ref{tab:ourLMs}) \textit{all} to a completely new test set. This additional evaluation iteration, which we describe in Section~\ref{subsec:additional_eval}  of the Appendix, verifies the undesirable effect of sharing users between the data splits.\footnote{Relevant results are in Appendix Table~\ref{tab:model_additional_eval}.} For this reason, we strongly advise against sharing users across data splits for tweet-level tasks \textit{even} if the overall dataset involves several thousand users. 

\begin{table}[h]
\centering
\begin{adjustbox}{width=\columnwidth}
\begin{tabular}{l|ll|ll|ll}
\hline
\multicolumn{1}{c|}{\textbf{Setting}} & \multicolumn{2}{c|}{\textbf{City}} & \multicolumn{2}{c|}{\textbf{State}} & \multicolumn{2}{c}{\textbf{Country}} \\ \hline
\multicolumn{1}{c|}{\textbf{Eval Metric}} & \textbf{acc} & \textbf{F1} & \textbf{acc} & \textbf{F1} & \textbf{acc} & \textbf{F1} \\ \hline
mBERT-wide & 4.58 & 3.18 & 7.88 & 4.05 & 35.39	 & 24.73\\
MARBERT-wide & \textbf{5.06} &	\textbf{3.59} & \textbf{8.92} &	\textbf{4.77} & \textbf{39.49} &	\textbf{28.77} \\ \hdashline
mBERT-medium & 6.39 &	4.92 & 9.33 &	6.11 & 	38.48 &	28.16\\
MARBERT-medium & \textbf{7.07} &	\textbf{5.81} & \textbf{10.53} &	\textbf{7.22} & \textbf{42.51} &	\textbf{32.37}\\ \hdashline

mBERT-narw & 9.60 & 6.63 &	12.08 &	8.55 &	51.32 &	33.87\\
MARBERT-narw & \textbf{11.66} &	\textbf{8.12} & \textbf{14.62} & \textbf{10.09} & \textbf{57.27} &	\textbf{39.93}\\ \hline  
MTL-MARBERT-narw (DiaGloss) & 11.76 &	8.35 & 14.37 &	10.23 &	56.74	& 39.68 \\
MTL-MARBERT-narw (CodSw) & 11.59	& 8.33 &	15.45	& 10.27	&	57.78	& 40.32 \\ 
MTL-MARBERT-narw (CSD) & \textbf{11.78}	& \textbf{8.44}	&	\textbf{15.25}	& \textbf{10.34}	& \textbf{57.75}	& \textbf{40.46}\\ 
\hline 
\end{tabular}%
\end{adjustbox}
\caption{Performance on TEST sets with unique users (i.e., users with no data in TRAIN). Setting names are suffixed to each model. \textbf{CSD:} DiaGloss+CodSw.}
\label{tab:model_generalization}
\end{table}



%% file: msa_impact.tex
\vspace*{-2mm}
\section{Impact of MSA}\label{sec:msa_impact}
\vspace*{-2mm}
Our efforts thus far focused on teasing apart tweets regardless of their (degree of) dialectness. Our dataset comprises posts either solely in MSA or in MSA mixed with dialectal content. Since MSA is shared across different regions, filtering it out is likely to enhance system ability to distinguish posts across our 3 classification tasks. We test (and confirm) this hypothesis by removing MSA from both TRAIN and TEST in our \textit{narrow} setting data (from Section~\ref{sec:model_gen}) and fine-tuning MARBERT on the resulting (`dialectal') data only.\footnote{To filter out MSA, we apply an in-house MSA vs. dialect classifier ($acc=89.1\%, F_1=88.6\%$) on the data, and remove tweets predicted as MSA. More information about the MSA vs. dialect model is in Section~\ref{subsec:mas_da_model} of the Appendix. We cast more extensive investigation of the interaction between dialects and MSA vis-a-vis our classification tasks, including based on manually-filtered MSA, as future research.  
} As Table~\ref{tab:MSA-DA} shows, this procedure results in higher performance across the 3 classification tasks.\footnote{MARBERT is significantly better than mBERT with $p<0.03$ for city, $p<0.01$ for state, and $p<0.0004$ for country.} For micro-dialects, performance is at $14.09\%$ acc. and $9.87\%$ F$_1$. Again, this performance is much better ($3.3\times$ better acc and $75.9\times$ better F$_1$) than the majority class baseline (city of Abu Dhabi, UAE, in 2 of our 3 runs). 

\begin{table}[h]
\centering
\begin{adjustbox}{width=\columnwidth}
\begin{tabular}{l|ll|ll|ll}
\hline
\multicolumn{1}{c|}{\textbf{Setting}} & \multicolumn{2}{c|}{\textbf{City}} & \multicolumn{2}{c|}{\textbf{State}} & \multicolumn{2}{c}{\textbf{Country}} \\ \hline
\multicolumn{1}{c|}{\textbf{Eval Metric}} & \textbf{acc} & \textbf{F1} & \textbf{acc} & \textbf{F1} & \textbf{acc} & \textbf{F1} \\ \hline
mBERT-narrow-DA & 11.72 & 8.45 & 15.41	& 10.95  & 	 69.30 & 45.46  \\
MARBERT-narrow-DA & \textbf{14.09} &	\textbf{9.87} & \textbf{17.37} &	\textbf{12.90} & \textbf{75.35} &	\textbf{51.03} \\ 
\hline 
\end{tabular}
\end{adjustbox}
\caption{Performance on predicted dialectal data.}
\label{tab:MSA-DA}
\end{table}

%% file: discussion.tex
\vspace*{-5mm}
\section{Discussion}\label{sec:discuss}
\vspace*{-2mm}
As we showed in Sections~\ref{sec:model_gen} and \ref{sec:msa_impact}, our models are able to predict variation at the city-level significantly better than \textit{all} competitive baselines. This is the case even when we do not remove MSA content, but better results are acquired after removing it. A question arises as to whether the models are indeed capturing micro-linguistic variation between the different cities, or simply depending on different topical and named entity distributions in the data. To answer this question, we visualize attention in $\sim250$ examples from our TEST set using one of our MARBERT-narrow models reported in Table~\ref{tab:model_generalization}.\footnote{Namely, we use the model fine-tuned in split B in Table~\ref{tab:model_gen_data}.} Our analysis reveals that the model \textit{does} capture micro-dialectal variation. We provide two example visualizations in Section~\ref{sec:append:discuss} in the Appendix demonstrating the model's micro-dialectal predictive power. Still, we also observe that the model makes use of especially names of places. For this reason, we believe future research should control for topical and named entity cues in the data.  

%% file: comparisons.tex
\section{Comparisons and Impact}\label{sec:comp}
\vspace*{-2mm}
\textbf{Comparisons to Other Dialect Models.} In absence of similar-spirited nuanced language models, we compare our work to existing models trained at the country level. These include the tweet-level 4-country SHAMI ~\cite{qwaider2018shami} which we split into TRAIN (80\%), DEV (10\%), and TEST (10\%) for our experiments, thus using less training data than ~\citet{qwaider2018shami} (who use cross-validation). We also compare to~\cite{zhang2019no}, the winning team in the the 21-country MADAR Shared Task-2~\cite{bouamor2019madar}. Note that the shared task targets user-level dialect based on a collection of tweets, which our models are not designed to directly predict (since we rather take a single tweet input, making our task harder).~\footnote{We follow~\cite{zhang2019no} in assigning a user-level label based on message-level majority class.} For the purpose, we train two models, one on MADAR data (shared tasks 1 and 2) and another on our Micro-Ara+MADAR data. We also develop models using the 17-country Arap-Tweet~\cite{zaghouani2018arap}, noting that authors did not preform classification on their data and so we include a unidirectional 1-layered GRU, with 500 units as a baseline for Arap-Tweet. 
Note that we do not report on the dataset described in ~\citet{mageedYouTweet2018} since it is automatically labeled, and so is \textit{noisy}. We also do not compare to the dataset in ~\citet{salameh2018fine} since it is small, \textit{not naturally occurring} (only 2,000 translated sentences per class), and the authors have already reported linear classifiers outperforming a deep learning model due to small data size. As Table~\ref{tab:marbert-external} shows, our models achieve new SOTA on all three tasks with a significant margin. Our results on MADAR show that if we have up to 100 messages from a user, we can detect their MD at 80.69\% acc.




\begin{table}[]
\centering
\footnotesize
\begin{adjustbox}{width=\columnwidth}
\begin{tabular}{@{}lrlll@{}}
\hline
\multicolumn{1}{c}{\textbf{Dataset}}                                                                 & \multicolumn{1}{c}{\textbf{\#cls}}               & \multicolumn{1}{c}{\textbf{Model} }   & \textbf{acc} & \textbf{F1} \\ \hline
\multirow{3}{*}{Arap-TWT}                                                        & \multirow{3}{*}{17} & GRU-500           & 38.79        & 39.17       \\
                                                                                 &                     & mBERT             & 54.67        & 55.07       \\
                                                                                 &                     & MARBERT          & \textbf{57.00}       &\textbf{57.21}     \\ \hdashline
\multirow{3}{*}{SHAMI}                                                           & \multirow{3}{*}{4}  & \newcite{qwaider2018shami}  & 70.00        & 71.00       \\
                                                                                 &                     & mBERT             & 86.07        & 85.46       \\
                                                                                 &                     & MARBERT          & \textbf{91.20}       &\textbf{87.70}      \\ \hdashline
\multirow{4}{*}{\begin{tabular}[c]{@{}l@{}}MADAR\\  (User-level)\end{tabular}} & \multirow{4}{*}{21} & \newcite{zhang2019no}          & 77.40        & 71.70       \\
                                                                                 &                     & mBERT (MST)       & 76.40        & 68.47       \\
                                                                                 &                     & MARBERT (MST)    & 76.39        & 70.61       \\
                                                                                 &                     & MARBERT (Micro-Ara+MST) & \textbf{80.69}       & \textbf{74.45}       \\ \hline
\end{tabular}
\end{adjustbox}
\caption{Results on external data. \textbf{MST:} MADAR task 1 and 2. MARBERT (ours) sets new SOTA on \textit{all} tasks.} 
\end{table} 

\textbf{Impact on External Tasks:}
We further demonstrate the impact of our newly-developed model, MARBERT, by fine-tuning it on a range of text classification tasks. These involve 4 sentiment analysis datasets: ArSAS~\cite{elmadany2018arsas}, ASTD~\cite{nabil2015astd}, SemEval-2017 task 4-A benchmark dataset~\cite{rosenthal2017semeval}, and Arabic sentiment collection (ASC) in ~\newcite{mageed2020aranet}; and a benchmark for offensive language (OFF) from OSACT4 Arabic Offensive Language Detection Shared Task~\cite{mubarak2020overview}. More information about each dataset can be found in the respective sources. We compare to the SOTA on each dataset, using the same metrics for each respective systems: For ArSAS, ASTD, and SemEval, we use $F_1^{PN}$.~\footnote{$F_1^{PN}$ was defined by SemEval-2017 as the $macro ~ F_1$ over the positive and negative classes only while neglecting the neutral class.} And for OFF and ASC, we use $macro ~ F_1$. As Table~\ref{tab:marbert-external} shows, our models set new SOTA on \textit{all} 5 datasets.


\begin{table}[h]
\centering
\begin{adjustbox}{width=\columnwidth}
\footnotesize
\begin{tabular}{l|ccccc}
\hline
               & ArSAS & ASTD & SemEv  & ASC & OFF\\
\hline
~\newcite{farha2019mazajak} (Mazaj)     & 90.00  & 72.00 & 63.00  &  --- &  --- \\
~\newcite{obeid2020camel}   (mBERT)      & 89.00  & 66.00 & 60.00   & --- &  --- \\ 
~\newcite{obeid2020camel}  (AraBERT)    & 92.00  & 73.00 & 69.00   & ---  &  --- \\
~\newcite{hassan2020alt} (AraBERT)    &  ---  &  ---    & ---    &      --- &  90.51  \\ 
~\newcite{mageed2020aranet} (mBRT)  & ---        & ---          & ---  & 76.67   &  ---      \\ 
\hline
\textbf{MARBERT}   (Ours)    & \textbf{92.50}  & \textbf{78.50} & \textbf{70.50}  &  \textbf{90.86} &  \textbf{91.47} \\

\hline
\end{tabular}%
\end{adjustbox}
\caption{Evaluation of MARBERT on external tasks.}
\label{tab:marbert-external}
\end{table}


%% file: lit.tex
\vspace*{-3mm}
\section{Related Work}\label{sec:lit}
\vspace*{-2mm}

\textbf{Dialectal Arabic Data and Models.} Much of the early work on Arabic varieties focused on collecting data for main varieties such as Egyptian and Levantine \cite{diab2010colaba,elfardy2012simplified,al2012yadac,sadat2014automatic,zaidan2011arabic}. Many works developed models for detecting 2-3 dialects \cite{elfardy2013sentence,zaidan2011arabic,zaidan2014arabic,cotterell2014multi}. Larger datasets, mainly based on Twitter, were recently introduced~\cite{mubarak2014using,mageedYouTweet2018,zaghouani2018arap,bouamor2019madar}. Our dataset is orders of magnitude larger than other datasets, more balanced, and more diverse. It is also, by far, the most fine-grained. 

\textbf{Geolocation, Variation, and MTL.} Research on geolocation is also relevant, whether based on text~\cite{roller2012supervised,graham2014world,han2016twitter,do2018twitter}, user profile~\cite{han2013stacking}, or network-based methods~\cite{miura2017unifying,ebrahimi2018unified}. Models exploiting network information, however, do not scale well to larger datasets~\cite{rahimi2015twitter}. ~\cite{eisenstein2012mapping} exploit geotagged Twitter data to study how words spread geographically.~\cite{bamman2014distributed} uses representations based on geolocation to improve semantic similarity.~\newcite{dunn2019global} studies syntactic variations in 7 languages based on geolocated data.
~\newcite{hovy2020visualizing} visualizes regional variation across Europe using Twitter. ~\newcite{dunn2020mapping} find that Twitter data are representative of actual population. MTL has been successfully applied to many NLP problems, including MT and syntactic parsing~\cite{luong2015multi}, sequence labeling~\cite{sogaard2016deep,rei2017semi}, and text classification~\cite{liu2016recurrent}.

%% file: conc.tex
\section{Conclusion}\label{sec:conc}
\vspace*{-2mm}
We introduced the novel task of MDI and offered a large-scale, manually-labeled dataset covering 319 city-based Arabic micro-varieties. We also introduced several novel MTL scenarios for modeling MDs including at hierarchical levels, and with linguistically-motivated auxiliary tasks inspired by diaglossic and code-switching environments. We have also exploited our own data to train MARBERT, a very large and powerful masked language model covering all Arabic varieties. Our models establish new SOTA on a wide range of tasks, thereby demonstrating their value. Ultimately, we hope our work can open up new horizons for studying MDs in various languages.

\section*{Acknowledgment}
We thank the anonymous reviewers for their valuable feedback, and Hassan Alhuzali for initial help with data. We gratefully acknowledge support from the Natural Sciences and Engineering Research Council of Canada, the Social Sciences Research Council of Canada, Compute Canada (\url{www.computecanada.ca}), and UBC ARC--Sockeye (\url{https://doi.org/10.14288/SOCKEYE}).

%% file: appendix.tex
\clearpage

\twocolumn[{%
 \centering
 \Large \textbf{Toward Micro-Dialect Identification in Diaglossic and Code-Switched Environments}\\[1em]
 
  \large Muhammad Abdul-Mageed  ~~~Chiyu Zhang  ~~~AbdelRahim Elmadany 
~~~Lyle Ungar$^\dagger$  \\ 
  Natural Language Processing Lab,  University of British Columbia \\
  $^\dagger$Computer and Information Science, University of Pennsylvania \\
  \tt\{muhammad.mageed,a.elmadany\}@ubc.ca \tt chiyuzh@mail.ubc.ca \\$^\dagger$\tt ungar@cis.upenn.edu \\[3em]
}]


\clearpage
\appendixpage
\addappheadtotoc
\counterwithin{figure}{section}
\counterwithin{table}{section}

\section{Data Acquisition and Labeling}\label{sec:append:data_acq}
\subsection{Correction of City and State Tags}\label{subsec:append:city_tag}
\textbf{City-Level.} Investigating examples of the geolocated data, we observed geopy made some mistakes. To solve the issue, we decided to manually verify the information returned from geopy on all the 705 assumed `cities'. For this purpose of manual verification, we use Wikipedia, Google maps, and web search as sources of information while checking city names. We found that geopy made mistakes in 7 cases as a result of misspelled city names in the queries we sent (as coming from user profiles). We also found that 44 cases were not assigned the correct city name as the first `solution'. Geopy provided us with a maximum of 7 solutions for a query, with best solutions sometimes being names of hamlets, villages, etc., rather than cities. In many cases, we found the correct solution to fall between the $2nd$ and $4th$ solutions. A third problem was that some city names (as coming from user profiles) were written in non-Arabic (e.g., English or French). We solved this issue by requiring geopy to also return the English version of a city name, and \textit{exclusively} using that English version. Ultimately, we acquired a total of 646 cities.

\textbf{State-Level.} 
As explained, geopy returned to us a total of 235 states/provinces that correspond to the 646 identified (manually fixed) cities. We also manually verified all the state names and their correspondence to the cities and countries. We found no issues with state tags. 

\subsection{Validation of User Location}\label{subsec:append:user-loc}

We trained the two annotators and instructed them to examine the profile information of each user on Twitter, providing a link to the profile. We asked them to consider various sources of information as a basis for their decisions, including (1) the profile picture, (2) profile textual description (including user-provided location), (3) the actual name of the user (if available), (4) at least 10 tweets, (5) the followers and followees of the user, and (5) user's network behavior such as the `likes'. 

Each annotator was responsible for $\sim 50\%$ of the usernames and was given a random sample of 100~\footnote{But we note that some cities had less than 100 users.} users for each city along with the Twitter handles and the automatically assigned \textit{city} and \textit{country} labels. We asked the users to label the first 10 accounts in each city, and only add more if the city proves specially challenging (as we observed to be the case in a pilot analysis of a few cities). Annotators ended up labeling a total of 4,953 accounts ($\sim 11.88$ users per city), of whom 4,012 users were verified for country and 3,085 for \textit{both} country and city locations.  We found that 81.00\% of geopy tags for country are correct, but only 62.29\% for city (which reduced our final city count to 319). As a final sanity check, a third annotator reviewed the labels for a random sample of 20 users from each annotator and agreed fully.

\section{Datasets}\label{sec:append:data}

\begin{table*}[!ht]
\centering
\footnotesize 
\begin{tabular}{
>{}l 
>{}c |r|rrr|rr}
\hline
\multicolumn{2}{c|}{\textbf{Countries}} &                                    & \multicolumn{3}{c|}{\textbf{\#Tweets}}                                                                                  &                                     &                                    \\ \cline{1-2} \cline{4-6}
\textbf{Name}                   & \textbf{Code}                  & {\textbf{\#Users}} & \textbf{Collected}   & \textbf{Retweets} & \textbf{Normalized}  & {\textbf{\#States}} & {\textbf{\#Cities}} \\ \hline
Algeria                         & dz                             & 1,960                                                      & 3,939,411                                    & 2,889,447                                         & 2,324,099                                    & 47                                                          & 200                                                         \\ 
Bahrain                         & bh                             & 1,080                                                      & 2,801,399                                    & 1,681,337                                         & 1,385,533                                    & 4                                                           & 4                                                           \\ 
Djibouti                        & dj                             & 6                                                          & 11,901                                       & 9,173                                             & 8,790                                        & 1                                                           & 1                                                           \\ 
Egypt                           & eg                             & 42,858                                                     & 92,804,863                                   & 61,264,656                                        & 47,463,301                                   & 27                                                          & 56                                                          \\ 
Iraq                            & iq                             & 4,624                                                      & 7,514,750                                    & 4,922,553                                         & 4,318,523                                    & 18                                                          & 62                                                          \\ 
Jordan                          & jo                             & 3,806                                                      & 7,796,794                                    & 5,416,413                                         & 4,209,815                                    & 4                                                           & 5                                                           \\ 
KSA                             & sa                             & 136,455                                                    & 297,264,647                                  & 177,751,985                                       & 165,036,420                                  & 11                                                          & 31                                                          \\ 
Kuwait                          & kw                             & 4,466                                                      & 11,461,531                                   & 7,984,758                                         & 6,628,689                                    & 4                                                           & 14                                                          \\ 
Lebanon                         & lb                             & 1,364                                                      & 3,036,432                                    & 1,893,089                                         & 1,160,167                                    & 6                                                           & 19                                                          \\ 
Libya                           & ly                             & 2,083                                                      & 4,227,802                                    & 3,109,355                                         & 2,655,180                                    & 21                                                          & 32                                                          \\ 
Mauritania                      & mr                             & 102                                                        & 209,131                                      & 148,261                                           & 129,919                                      & 4                                                           & 4                                                           \\ 
Morocco                         & ma                             & 1,729                                                      & 3,407,741                                    & 2,644,733                                         & 1,815,947                                    & 17                                                          & 117                                                         \\ 
Oman                            & om                             & 4,260                                                      & 8,139,374                                    & 4,866,813                                         & 4,259,780                                    & 8                                                           & 17                                                          \\ 
Palestine                       & ps                             & 2,854                                                      & 6,004,791                                    & 4,820,335                                         & 4,263,491                                    & 2                                                           & 12                                                          \\ 
Qatar                           & qa                             & 5,047                                                      & 11,824,490                                   & 7,891,425                                         & 6,867,304                                    & 2                                                           & 2                                                           \\ 
Somalia                         & so                             & 78                                                         & 168,136                                      & 131,944                                           & 104,946                                      & 8                                                           & 9                                                           \\
Sudan                           & sd                             & 1,162                                                      & 2,348,325                                    & 1,522,274                                         & 1,171,866                                    & 14                                                          & 27                                                          \\ 
Syria                           & sy                             & 1,630                                                      & 2,992,106                                    & 2,184,715                                         & 1,889,455                                    & 12                                                          & 19                                                          \\ 
Tunisia                         & tn                             & 227                                                        & 460,268                                      & 362,806                                           & 239,769                                      & 10                                                          & 10                                                          \\ 
UAE                             & ae                             & 14,923                                                     & 36,121,319                                   & 23,309,788                                        & 18,484,296                                   & 7                                                           & 15                                                          \\ 
Yemen                           & ye                             & 2,391                                                      & 4,783,144                                    & 3,368,262                                         & 3,013,517                                    & 8                                                           & 8                                                           \\ \hline
\multicolumn{2}{c|}{\textbf{Total}}     & \textbf{233,105}                   & \textbf{507,318,355} & \textbf{318,174,122}      & \textbf{277,430,807} & \textbf{235}                        & \textbf{664}                        \\ \hline
\end{tabular}
\center 
\caption{Statistics of our data representing 233,105 users from 664 cities and 21 countries. We process more than half a billion tweets, from a larger pool of $\sim$6 billion tweets, to acquire our final dataset. Note that the number of states and cities is further reduced after our manual user verification. Eventually, we acquire data for 319 cities, belonging to 192. The data represent all 21 Arab countries.}
\label{tab:data_stats}
\end{table*}
\begin{figure}[h]
\center 
\frame{\includegraphics[width=\linewidth]{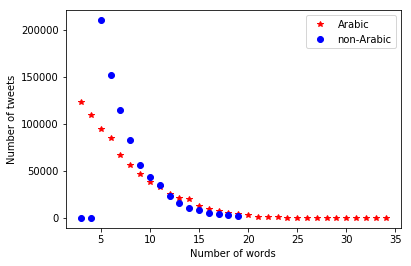}}
\caption{Word frequency distribution in our CS-21 (code-switching) dataset.}
\label{fig:cs_freq}
\end{figure}
\begin{figure*}[h]
\center 
\frame{\includegraphics[width=\columnwidth*2]{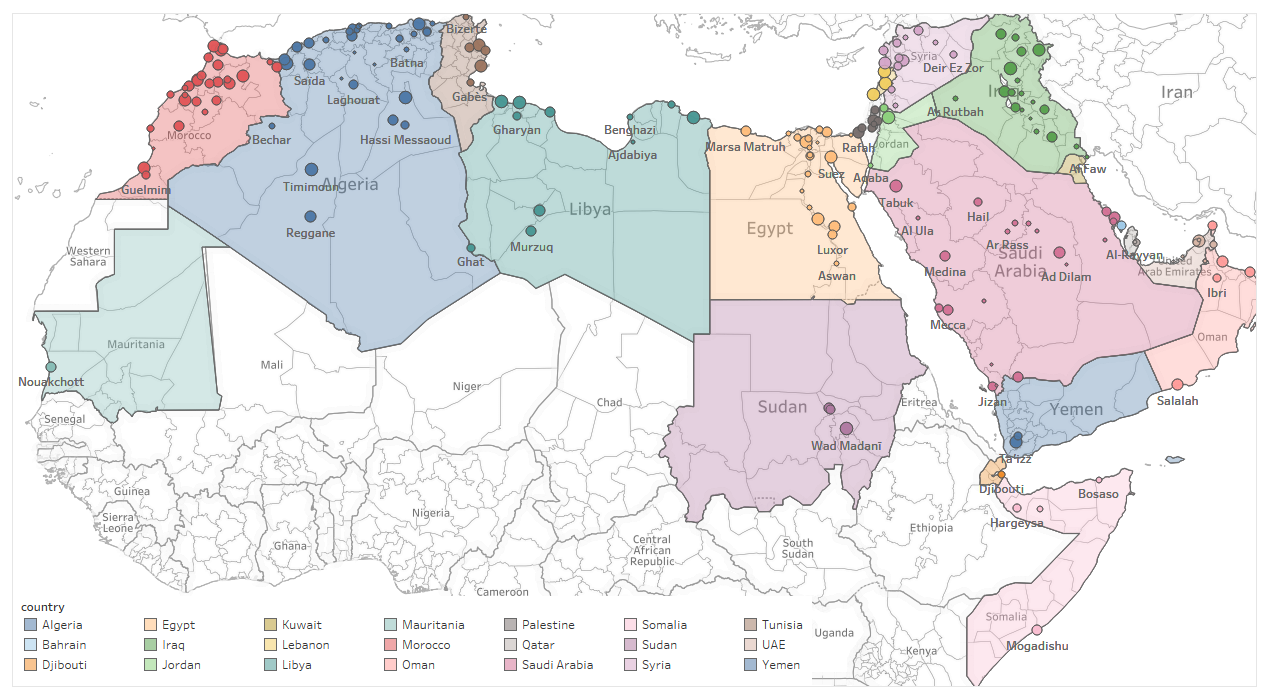}}
\caption{A bigger-sized map of all 21 Arab countries. States are demarcated in thin black lines within each country. A total of 319 cities (from our user location validation study, in colored circles) are overlayed within corresponding countries.}
\label{fig:all_loc_big}
\end{figure*}


\begin{table}[H]
\footnotesize 
\begin{tabular}{lccr}
\hline
\textbf{Country} & \textbf{\%vld\_cntry} & \textbf{\%vld\_city} & \textbf{\#tweets} \\ \hline
Algeria & 77.49\ & 69.74\ & 185,854 \\
Bahrain & 83.95\ & 39.51\ & 25,495 \\
Djibouti & 68.42\ & 68.42\ & 3,939 \\
Egypt & 92.66\ & 64.02\ & 463,695 \\
Iraq & 51.50\ & 37.61\ & 59,287 \\
Jordan & 83.61\ & 54.10\ & 17,958 \\
KSA & 96.37\ & 62.88\ & 353,057 \\
Kuwait & 84.30\ & 34.88\ & 65,036 \\
Lebanon & 92.42\ & 56.06\ & 37,273 \\
Libya & 75.48\ & 72.03\ & 128,152 \\
Maurit. & 45.00\ & 35.00\ & 3,244 \\
Morocco & 75.59\ & 62.42\ & 140,341 \\
Oman & 90.25\ & 77.97\ & 108,846 \\
Palestine & 87.50\ & 82.35\ & 87,446 \\
Qatar & 85.00\ & 77.50\ & 29,445 \\
Somalia & 52.73\ & 45.45\ & 9,640 \\
Sudan & 56.88\ & 41.28\ & 23,642 \\
Syria & 76.28\ & 71.63\ & 79,649 \\
Tunisia & 78.95\ & 75.94\ & 26,300 \\
UAE & 85.31\ & 82.49\ & 129,264 \\
Yemen & 72.41\ & 56.32\ & 47,450 \\ \hline
\textbf{Avg/Total} & 81.00\ & 62.29\ & 2,025,013 \\\hline
\end{tabular}
\caption{Our gold data, from manually verified users.}
\label{tab:gold_data}
\end{table}

\begin{table}[!ht]
\centering
\footnotesize 
\begin{tabular}{lrrr} \hline
\textbf{Country} & \textbf{TRAIN} & \textbf{DEV} & \textbf{TEST} \\ \hline
Algeria & 100,000 & 18,700 & 18,572 \\
Bahrain & 20,387 & 2,556 & 2,552 \\
Djibouti & 3,158 & 408 & 373 \\
Egypt & 100,000 & 46,136 & 46,325 \\
Iraq & 47,395 & 5,903 & 5,989 \\
Jordan & 14,413 & 1,826 & 1,719 \\
KSA & 100,000 & 35,312 & 35,106 \\
Kuwait & 52,127 & 6,416 & 6,493 \\
Lebanon & 29,821 & 3,641 & 3,811 \\
Libya & 100,000 & 12,847 & 12,803 \\
Maurit. & 2,579 & 338 & 327 \\
Morocco & 100,000 & 14,118 & 13,862 \\
Oman & 87,048 & 10,807 & 10,991 \\
Palestine & 69,834 & 8,668 & 8,944 \\
Qatar & 23,624 & 2,968 & 2,853 \\
Somalia & 7,678 & 1,023 & 939 \\
Sudan & 18,929 & 2,334 & 2,379 \\
Syria & 63,668 & 7,987 & 7,994 \\
Tunisia & 21,164 & 2,599 & 2,537 \\
UAE & 100,000 & 13,089 & 12,768 \\
Yemen & 37,886 & 4,833 & 4,731 \\\hline
Total & 1,099,711 & 202,509 & 202,068 \\ \hline
\end{tabular}
\caption{Distribution of classes in our data splits.}
\label{tab:splits}
\end{table}
\newpage
\newpage
\section{Models}\label{sec:append:models}

\subsection{Single Task BiGRUs (Second Baseline)}\label{subsec:single_task_BiGrus}

As mentioned in Section~\ref{subsec:baselines}, our a \textit{second baseline} (Baseline II), is comprised of 3 independent networks (each for one of the 3 tasks) using the same architecture and model capacity. Each network has 3 hidden BiGRU layers,~\footnote{We also ran single-task networks with 4 hidden layers, but we find them to overfit quickly even when we regularize with dropout at $0.7$ on all layers.} with 1,000 units each (500 units from left to right and 500 units from right to left). We add dot-product attention \textit{only} to the third hidden layer. We trim each sequence at 50 words,~\footnote{In initial experiments, we found a maximum sequence of $30$ words to perform slightly worse.} and use a batch size of 8. Each word in the input sequence is represented as a vector of 300 dimensions that are learned directly from the data. Word vectors weights $W$ are initialized with a normal distribution, with $\mu=0$, and $\sigma=0.05$, i.e., $W \sim N(0,0.05)$. For optimization, we use Adam~\cite{kingma2014adam} with a fixed learning rate of $1e-3$. For regularization, we use dropout~\cite{srivastava2014dropout} with a value of $0.5$ on each of the 3 hidden layers.

\subsection{Distill BERT}\label{subsec:distilbert}
We distill mBERT knowledge in out HA-MTL BiGRUs. In other words, we use the output of the mBERT logit layer as input to our city-first and country-first HA-MTL BiGRUs to optimize a mean-squared error objective function, but not a cross-entropy function (following equation 3 in ~\newcite{tang2019distilling}).~\footnote{The network architecture of the HA-MTL BiGRU is otherwise similar as before, but we train them for 20 epochs rather than 15.} As Table~\ref{tab:appendix_test_results}, both of these networks (HA-MTL-Dist-city $1^{st}$ and HA-MTL-Dist-country $1^{st}$ in the table) acquire sizeable improvements over the equivalent, non-distilled BiGRUs. Although these distillation models are still less than BERT, the goal behind them is to yield as closer-as-possible performance to BERT albeit with a smaller network that can be deployed in machines with limited capacity and offer quicker inference. Concretely, a HA-MTL-BiGRU model learns the 3 tasks of city, state, and country together compared to the single task BERT where 3 different models are needed for these 3 tasks. In terms of the number of parameters, this means the multi-task BiGRU distillation model has \textit{$11.6\times$ fewer parameters}. HA-MTL-BiGRU is also \textit{1.7 times faster} at inference.~\footnote{We perform model inference on the DEV set with a batch size of 128 on a single NVIDIA V100 GPU.}
\subsection{Multi-Task BiGRUs}
\begin{figure}[h!]
\centering
\frame{\includegraphics[width=\linewidth]{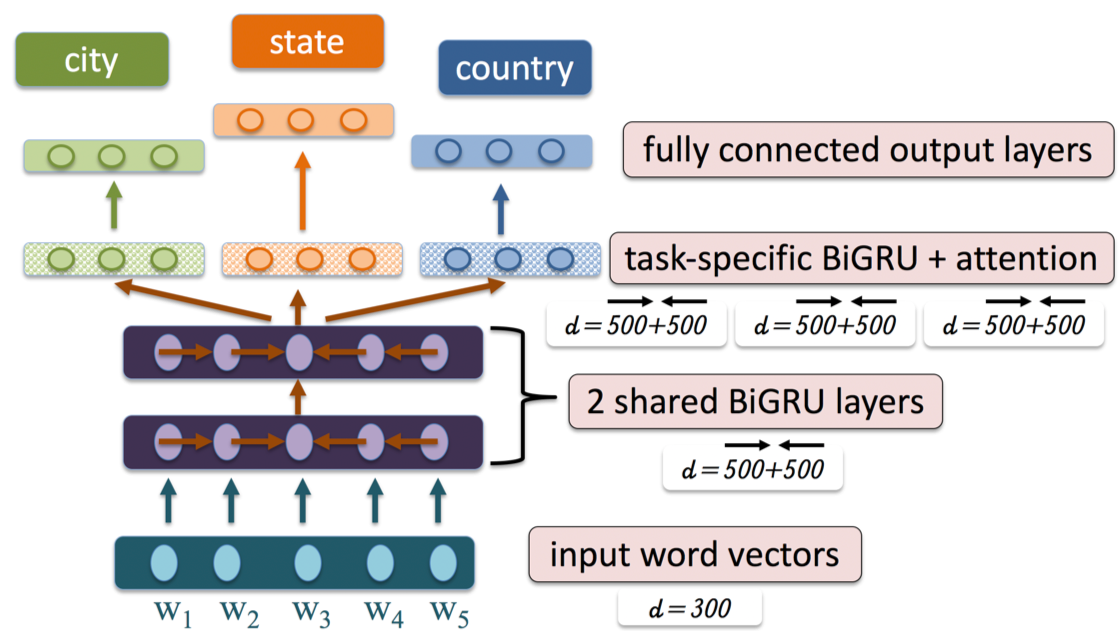}}
\caption{Illustration of MTL (spec-attn) network for city, state, and country. The three tasks share 2 hidden layers, with each task having its independent attention layer.}
\label{fig:mtl_t_spec_attn}
\end{figure}

\begin{figure}[h!]
\centering
\frame{\includegraphics[width=\linewidth]{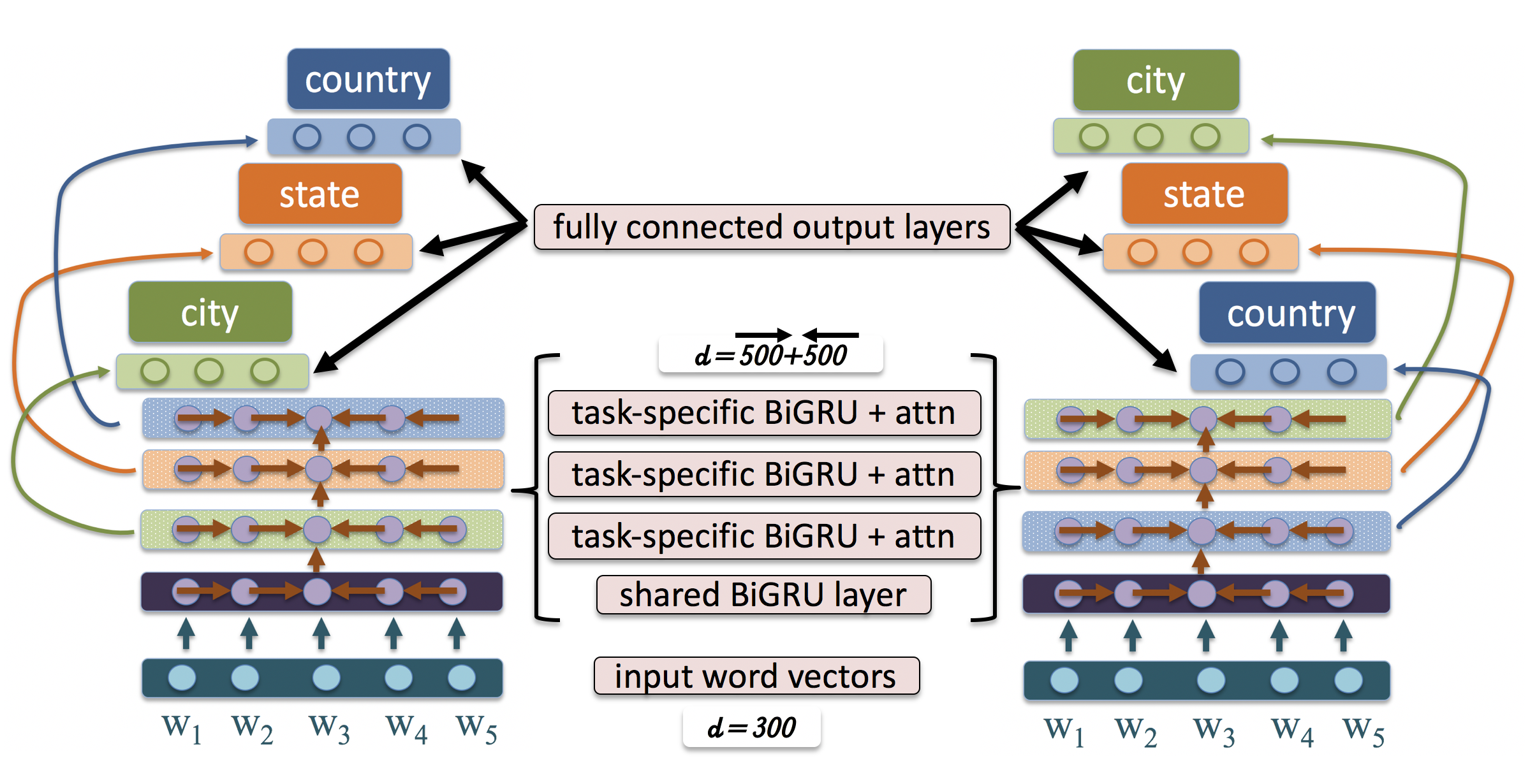}}
\caption{Hierarchical Attention MTL of city, state, and country. All models share one BiGRU layer of 1,000 units. Layers 2-4 are also BiGRU layers, with dot-product attention. \textbf{Left:} City network supervised at layer 2, state at layer 3, and country at layer 4. \textbf{Right:} Supervision is reversed from left network.}\label{fig:mtl2b}
\end{figure}

\begin{table*}[h]
\centering
\footnotesize
\begin{tabular}{l|rr|rr|rr}
\hline
\multicolumn{1}{c|}{\textbf{Setting}} & \multicolumn{2}{c|}{\textbf{City}} & \multicolumn{2}{c|}{\textbf{State}} & \multicolumn{2}{c}{\textbf{Country}} \\ \hline
\multicolumn{1}{c|}{\textbf{Eval Metric}} & \textbf{acc} & \textbf{F1} & \textbf{acc} & \textbf{F1} & \textbf{acc} & \textbf{F1} \\ \hline
Baseline I & 1.31 & 0.01 & 3.11 & 0.03 & 9.19 & 0.80 \\
Baseline II  & 1.65 & 0.25 & 6.13 & 1.92 & 31.14 & 15.84 \\ \hline
MTL (common-attn) & 2.86 & 0.74 & 5.12 & 1.01 & 26.51 & 12.41 \\
MTL (spec-attn) & 2.40 & 0.68 & 4.60 & 0.90 & 27.04 & 10.98 \\
HA-MTL (city $1^{st}$) & \textbf{14.01} & \textbf{14.02} & \textbf{16.14} & \textbf{15.90} & \textbf{44.36} & 32.14 \\
HA-MTL (cntry $1^{st}$) & 13.23 &	13.06 & 15.84	& 15.40 & 44.17	& \textbf{32.37} \\\hline
mBERT(\newcite{devlin2018bert}) & \textbf{19.33} & \textbf{19.45} & \textbf{21.24} & \textbf{21.67} & \textbf{47.74} & \textbf{38.12}\\
HA-MTL-Dist (city $1^{st}$) & 15.69 & 15.95 & 18.33 & 18.72 & 46.37 & 36.27 \\
HA-MTL-Dist (cntry $1^{st}$) & 15.73	& 15.58   & 18.21	 & 18.46 & 46.34 &  36.13\\
\hdashline 
MTL-mBERT (DiaGloss)& \textbf{19.88} & \textbf{20.11} & 21.04 & \textbf{21.69} & 48.30 & \textbf{38.34}\\
MTL-mBERT (CodSw)& 19.47 & 19.86 & 20.76 & 21.47 & \textbf{48.61} & 38.20 \\ 
\hline 
AraBERT (\newcite{baly2020arabert}) &18.82 &	18.73 & 20.73 &20.87 & 48.33  & 38.09
\\ 
MARBERT (Ours) & \textbf{20.78} & \textbf{20.41} &\textbf{ 22.97} & \textbf{22.58} & \textbf{51.87} & \textbf{42.17}\\
\hdashline
MTL-MARBERT (DiaGloss)& 20.19 & \textbf{20.60} & \textbf{23.22} & 22.97 & 51.53 & 41.75\\
MTL-MARBERT (CodSw) & \textbf{20.77}  & 20.56  & 23.21 & \textbf{23.16} & 51.78 & \textbf{42.36} \\ 
MTL-MARBERT (CSD) & 20.76 & 20.23 & 23.18 & 23.15 & \textbf{52.17}& 42.27 \\ 
\hline 
\end{tabular}
\caption{Performance on TEST. Baseline I: majority in TRAIN. Baseline II: single task Attn-BiGRU. HA-MTL-Dist: BiGRU distilling of mBERT knowledge. MTL-mBERT/MARBERT (CodSw): Code-switching at ``country" level. MTL-MARBERT (CSD): Two auxiliary tasks, code-switching supervised at country level and diglossia.}
\label{tab:appendix_test_results}
\end{table*}
\begin{table*}[h!]
\centering
\footnotesize
\begin{tabular}{l|rr|rr|rr}
\hline
\multicolumn{1}{c|}{\textbf{Setting}} & \multicolumn{2}{c|}{\textbf{City}} & \multicolumn{2}{c|}{\textbf{State}} & \multicolumn{2}{c}{\textbf{Country}} \\ \hline
\multicolumn{1}{c|}{\textbf{Eval Metric}} & \textbf{acc} & \textbf{F1} & \textbf{acc} & \textbf{F1} & \textbf{acc} & \textbf{F1} \\ \hline
Baseline I & 1.31 & 0.01 & 3.11 &	0.03 &  9.19	& 0.80  \\ 
Baseline II  & 1.72 & 0.26 & 6.08 & 1.92 & 30.94 & 15.84 \\ \hline
MTL (common-attn) & 2.90 & 0.74 & 5.07 & 1.04 & 26.52 & 12.44 \\
MTL (spec-attn) & 2.48 & 0.70 & 4.57 & 0.90 & 27.04 & 10.98 \\
HA-MTL (city $1^{st}$) & \textbf{14.08} & \textbf{14.29} & \textbf{16.10} & \textbf{16.00} & \textbf{44.14} & \textbf{33.14} \\
HA-MTL (cntry $1^{st}$) & 13.31 &	13.20 & 15.91	& 15.78 & 44.06	& 32.79 \\\hline
mBERT & \textbf{19.56}	& \textbf{19.82}   & \textbf{21.20} \ & \textbf{21.67} \ & \textbf{47.57} \ & \textbf{38.30}\\
HA-MTL-Dist (city $1^{st}$) & 15.79 & 15.96 & 18.38 & 18.69 & 46.11 & 36.50 \\
HA-MTL-Dist (cntry $1^{st}$) & 15.85	& 15.76  & 18.28	 & 18.39 & 46.22 &  36.21\\
\hdashline 
MTL-mBERT (DiaGloss)& \textbf{19.92} &\textbf{20.57} & 20.89 & 21.58 & 48.16 & 38.43\\
MTL-mBERT (CodSw)& 19.56 & 20.25 & \textbf{21.09} & \textbf{21.92} & \textbf{48.57} & \textbf{38.66} \\ 
\hline 
\end{tabular}
\caption{Performance on DEV. Baseline I: majority in TRAIN. Baseline II: single task Attn-BiGRU. MTL-mBERT (CodSw): Code-switching at ``country" level. HA-MTL-Dist: BiGRU distilling of mBERT knowledge.}
\label{tab:appendix_dev_res}
\end{table*}
\begin{table*}[h!]
\centering
\footnotesize
\begin{tabular}{l|ll|ll|ll}
\hline
\multicolumn{1}{c|}{\textbf{Setting}} & \multicolumn{2}{c|}{\textbf{City}} & \multicolumn{2}{c|}{\textbf{State}} & \multicolumn{2}{c}{\textbf{Country}} \\ \hline
\multicolumn{1}{c|}{\textbf{Eval Metric}} & \textbf{acc} & \textbf{F1} & \textbf{acc} & \textbf{F1} & \textbf{acc} & \textbf{F1} \\ \hline
mBERT (\newcite{devlin2018bert}) & 19.56 & 19.82 & 21.20 & 21.67 & 47.57 & 38.30\\ 
AraBERT (\newcite{baly2020arabert}) &18.77 & 18.69 & 20.63 & 21.16 & 48.18  & 38.53
\\ \hline
MARBERT (Ours) & \textbf{20.81} & \textbf{20.29} & \textbf{23.05} & \textbf{22.92} & \textbf{51.73} & \textbf{42.59}\\
\hdashline
MTL-MARBERT (DiaGloss)& \textbf{20.86} & \textbf{20.87} & 23.15 & 23.17 & 51.34 & 42.07\\
MTL-MARBERT (CodSw) & 20.85  & 20.78  & 23.13 & 23.03 & 51.60 & 42.37 \\ 
MTL-MARBERT (CSD) & 20.76 & 20.36 & \textbf{23.23} & \textbf{23.24} & \textbf{52.06}& \textbf{42.52} \\ 
\hline 
\end{tabular}%
\caption{Performance on DEV. MTL-MARBERT (CSD): Two auxiliary tasks, code-switching supervised at country level and diglossia.}
\label{tab:ourLMs_appendix_dev}
\end{table*}
\section{Model Generalization}\label{append:model_gen}
\vspace*{-2mm}

\subsection{Different Data Splits}\label{subsec:narrow_med_wide}

\begin{figure}[h!]
\tiny  
  \centering
  \frame{\includegraphics[width=\columnwidth]{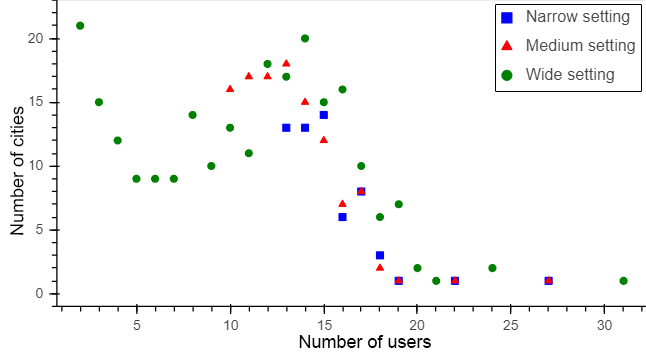}}
 
  \caption{Distribution of users over cities across our 3 model generalization settings.}
\label{fig:users_by_cities}
\end{figure}

\textbf{Narrow, Medium, and Wide Settings.} For \textbf{(1) narrow}, we select data from cities where we have at least 16 users, dividing users randomly into 3 in TEST and the rest (13 or more) in TRAIN. This gives us 61 cities, 48 states, and 11 countries. Our \textbf{(2) medium} setting is similar to \textit{narrow}, but we sample from cities with at least 13 users instead of 16. We use 3 users for TEST and the rest (10 users or more) for TRAIN. This results in 116 cities, 90 states, and 17 countries. \textbf{(3) Wide} has data from a single user from a given city in TEST and the rest of users from the same city in TRAIN. This setting allows more coverage (240 cities, 158 states, and all the 21 countries), at the cost of having as few as only two users for a given city in TRAIN. Figure~\ref{fig:users_by_cities} shows the distribution of users over the 3 data settings. In addition, Table~\ref{tab:model_gen_data} shows the data sizes of the TRAIN and TEST sets in each of the 3 runs, across each of the 3 settings. 

\subsection{Comparison of Models on a Completely New TEST Set.}\label{subsec:additional_eval}
\begin{table*}[h!]
\centering
\resizebox{\textwidth}{!}{%
\begin{tabular}{lrrrrrr|rrrrrr}
\hline
\multicolumn{1}{c}{}                            & \multicolumn{6}{c}{\textbf{DEV}}                                                                                                                                                                                                                                                                                              & \multicolumn{6}{c}{\textbf{TEST}}                                                                                                                                                                                                                                                                                             \\ \cline{2-13} 
\multicolumn{1}{c}{\multirow{-2}{*}{\textbf{Model}}} & \multicolumn{1}{c}{\textbf{acc}} & \multicolumn{1}{c}{\textbf{F1}} & \multicolumn{1}{c}{\textbf{acc@161}} & \multicolumn{1}{c}{\textbf{acc@80.5}} & \multicolumn{1}{c}{\textbf{mean(K)}} & \multicolumn{1}{c}{\textbf{median(K)}}& \multicolumn{1}{|c}{\textbf{acc}} & \multicolumn{1}{c}{\textbf{F1}} & \multicolumn{1}{c}{\textbf{acc@161}} & \multicolumn{1}{c}{\textbf{acc@80.5}} & \multicolumn{1}{c}{\textbf{mean(K)}} & \multicolumn{1}{c}{\textbf{median(K)}} \\ \hline
{MARB-319}                                       & 3.23                                           & 2.88                                          & 12.32                                                   & 7.07                                                     & 1,509.41                                               & 1,140.17                                                 & 3.12                                           & 2.53                                          & 15.11                                                   & 7.83                                                     & 1,446.83                                               & 1,022.29                                                 \\ 
{Wide}                                      & 3.90                                           & 3.70                                          & 12.71                                                   & 7.50                                                     & 1,438.68                                               & 1,070.05                                                 & 3.49                                           & 2.97                                          & 15.27                                                   & 7.91                                                     & 1,370.23                                               & 936.60                                                  \\ 
{Medium}                                    & 5.24                                           & 4.69                                          & 15.85                                                   & 9.70                                                     & 1,171.39                                               & 734.98                                                  & 5.43                                           & 4.56                                          & 19.91                                                   & 10.58                                                    & 1,151.77                                               & 652.06                                                  \\ 
{Narrow}                                      & 7.66                                           & 6.85                                          & 17.95                                                   & 12.58                                                    & 1,036.58                                               & 678.04                                                  & 6.56                                           & 5.49                                          & 20.69                                                   & 12.25                                                    & 1,052.54                                               & 624.09                                                  \\ \hline
\end{tabular}
}
 \caption{Evaluation of our MARBERT-based models on a new TEST set (GeoAra). \textbf{MARB-319:} our single task MARBERT model trained on 319 cities (reported on Table~\ref{tab:ourLMs}). \textbf{Narrow:} MARBERT-narrow, \textbf{Medium:} MARBERT-medium; \textbf{Wide:} MARBERT-wide.  \textbf{Acc@80.5:} Accuracy at 80.5 kilometers (=50 miles). \textbf{Acc@161:} Accuracy at 161 kilometers (=100 miles). \textbf{Mean(K):} Mean distance in kilometers. \textbf{Median(K):} Median distance in kilometers.}\label{tab:model_additional_eval}
\end{table*}

As mentioned in Section~\ref{sec:model_gen}, we evaluate our models from the narrow, medium and wide settings \textit{and} our single task MARBERT model (shown in Table~\ref{tab:ourLMs}) all on a completely new test set. This allows a more direct comparison between these models, including to test the impact of \textit{sharing users} across the various data splits (as is the case of single task MARBERT) or \textit{lack thereof} (as is the case for the narrow, medium and wide settings models). We now introduce GeoAra, our new evaluation dataset.

\textbf{GeoAra Dataset.}\label{subsec:geo_data}
GeoAra is a dataset of tweets with city labels from 20 Arab countries.~\footnote{These are the same countries as in our MicroAra, with the exception of Djibouti.} To build GeoAra, we run a crawler on each of the 319 cities in our gold data for a total of 10 month (Jan. 2019 - Oct. 2019). We acquire a total of $4.7$M tweets from all the cities. We collect Twitter user ids from users who posted consistently from a single location over the whole 10 months (n= 390,396), and crawl the timeline of $148$K users.~\footnote{We note that this is more conservative than previous geolocation works (e.g., ~\cite{han2012geolocation}) that take the majority class city of a user who posted 10 tweets as the label.} Note that MicroAra (our monolingual dataset) is collected in 2016 and 2017. This means GeoAra involves data from a period significantly different (more recent) than MicroAra (2 years later). We then \textit{only} keep users who posted at least 10 tweets. This leaves us with 101,960 users from 147 cities. From GeoAra, we create a DEV set from a random sample of 100K tweets (908 from users) and a TEST set from a random sample of 97,834 tweets (from 1,053 users).~\footnote{The two splits do not identically match since we also needed to create a specific TRAIN split from the same dataset. The TRAIN split is not part of the current work and so we leave it out.} We \textit{do not} share users between the TRAIN, DEV, and TEST splits. 

As Table~\ref{tab:model_additional_eval} shows, although all the 4 models degrade on GeoAra, single task MARBERT (MARB-319 in the table) suffers most. This further suggests, that this particular model has captured user-level knowledge that may have allowed it to perform much higher on the TEST set in Table~\ref{tab:ourLMs} than what it would if user data were not shared across the various splits. In addition, even though our \textit{narrow setting} model covers only 61 cities, it is the one that performs best on both the DEV and TEST GeoAra splits. This might be the case because this model is trained on the most number of users (at least 13 users for each city), which allows it to generalize well on these cities. An error analysis may reveal more information on performance of these particular models on GeoAra. We cast further investigation of this issue as future research.

\begin{table}[h!]
\centering
\footnotesize
\resizebox{0.5\textwidth}{!}{%
\begin{tabular}{lcrrrrr}
\hline
\multicolumn{1}{l}{}             & \textbf{split} & \multicolumn{1}{c}{\textbf{TRAIN}} & \multicolumn{1}{c}{\textbf{TEST}} & \multicolumn{1}{c}{\textbf{\#cities}} & \multicolumn{1}{c}{\textbf{\#states}} & \multicolumn{1}{c}{\textbf{\#countries}} \\ \hline
\multirow{3}{*}{{Narrow}} & {A}     & 444,838                                 & 87,540                                 & \multirow{3}{*}{61}                   & \multirow{3}{*}{48}                   & \multirow{3}{*}{11}                      \\
                                 & {B}     & 445,297                                 & 87,081                                 &                                       &                                       &                                          \\
                                 & {C}     & 453,451                                 & 78,927                                 &                                       &                                       &                                          \\ \hdashline
\multirow{3}{*}{{Medium}} & {A}     & 807,590                                 & 176,337                                & \multirow{3}{*}{116}                  & \multirow{3}{*}{90}                   & \multirow{3}{*}{17}                      \\
                                 & {B}     & 808,667                                 & 175,260                                &                                       &                                       &                                          \\
                                 & {C}     & 810,471                                 & 173,456                                &                                       &                                       &                                          \\ \hdashline
\multirow{3}{*}{{Wide}}   & {A}     & 1,308,640                               & 124,834                                & \multirow{3}{*}{240}                  & \multirow{3}{*}{158}                  & \multirow{3}{*}{21}                      \\
                                 & {B}     & 1,309,705                               & 123,769                                &                                       &                                       &                                          \\
                                 & {C}     & 1,305,098                               & 128,376                                &                                       &                                       &                                          \\ \hline
\end{tabular}\caption{TRAIN and TEST data sizes and label distribution (in city, state, and country) across the 3 splits for each of the \textit{narrow}, \textit{medium}, and \textit{wide} settings. }\label{tab:model_gen_data}
}
\end{table}

\begin{figure*}[h!]
\tiny  
  \centering
  \frame{\includegraphics[width=\textwidth]{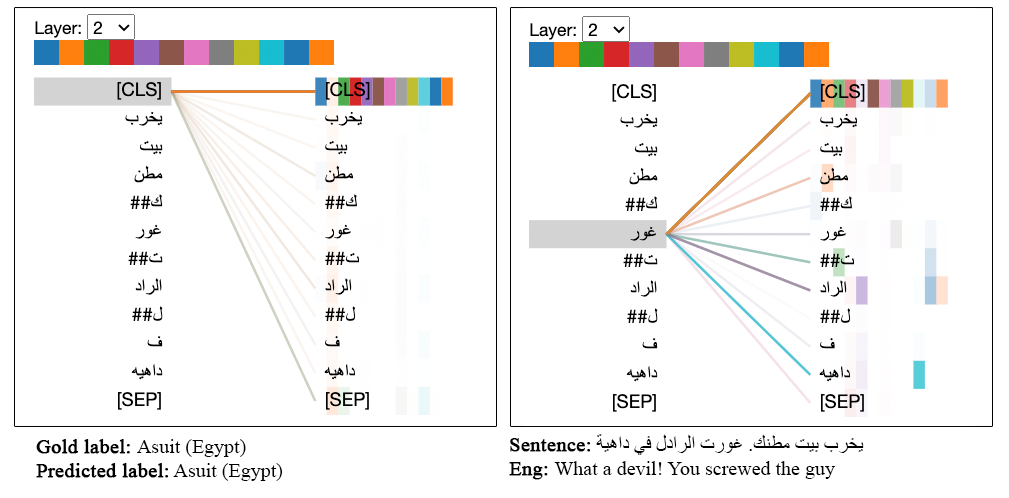}}
 
  \caption{An example sentence where the model correctly predicted the gold city (Asuit, Egypt), clearly laying attention on relevant micro-dialectal tokens.}
\label{fig:attn_ex1}
\end{figure*}

\begin{figure*}[h!]
\tiny  
  \centering
  \frame{\includegraphics[width=\textwidth]{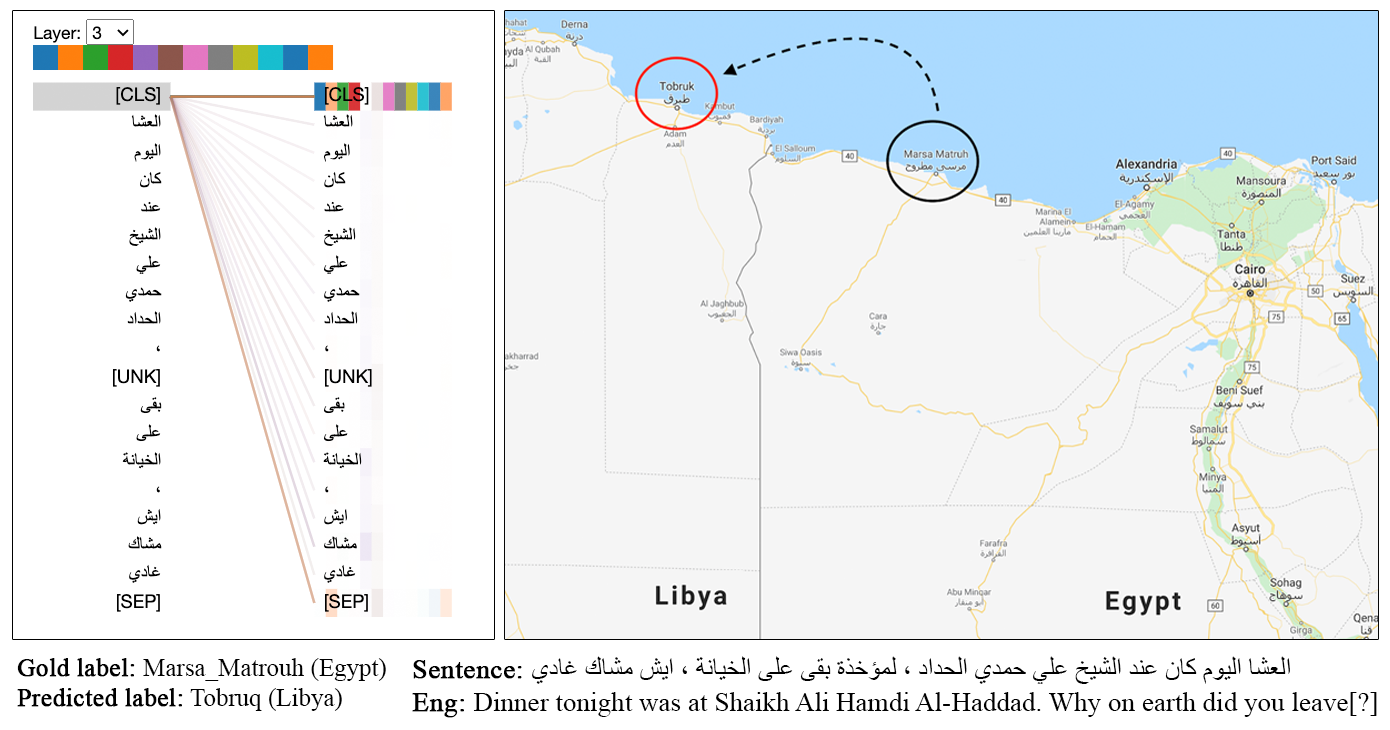}}
 
  \caption{An example of model error. The model confuses the city of Marsa Matrouh (Egypt, black circled) with that of Tobruq (Libya, red circled). The visualization illustrates how the model is capable of relating language across country boundaries, suggesting it does posses micro-dialectal predictive power.}
\label{fig:attn_ex2}
\end{figure*}

 

 
\begin{table}[h!]
\footnotesize
  \centering
\begin{tabular}{lllrr}
\hline
               & \multicolumn{2}{c}{\textbf{Dialect}}                                   & \multicolumn{2}{c}{\textbf{MSA}}                                       \\ \hline
\textbf{split} & \multicolumn{1}{c}{\textbf{TRAIN}} & \multicolumn{1}{c}{\textbf{TEST}} & \multicolumn{1}{c}{\textbf{TRAIN}} & \multicolumn{1}{c}{\textbf{TEST}} \\ \hline
A              & 218,231                            & 44,641                            & 226,607                            & 42,899                            \\
B              & 220,697                            & 42,175                            & 224,600                            & 44,906                            \\
C              & 225,737                            & 37,135                            & 227,714                            & 41,792                            \\ \hline
\end{tabular}
\caption{Distribution of MSA and DA over TRAIN and TEST of \textit{narrow} setting.}\label{tab:msa_da_narrow}
\end{table}

\subsection{MSA vs. Dialect Classifier}\label{subsec:mas_da_model}
As described in Section~\ref{sec:msa_impact}, we apply an MSA vs. dialect in-house classifier on our \textit{narrow} data setting, to remove MSA. Our in-house classifier fine-tuned MARBERT on the MSA-dialect TRAIN split described in~\newcite{elaraby-abdul-mageed-2018-deep}. This binary classifier performs at $89.1\%$ accuracy, and $88.6\%$ $F_1$ on ~\newcite{elaraby-abdul-mageed-2018-deep} MSA-dialect TEST set. Running this model on our \textit{narrow} setting data, gives us the TRAIN and TEST splits with predicted labels described in Table~\ref{tab:msa_da_narrow}. 

\section{Discussion}\label{sec:append:discuss}
As discussed in Section~\ref{sec:discuss}, we visualize attention in $\sim250$ examples from our TEST set using our MARBERT-narrow model fine-tuned in split B in Table~\ref{tab:model_gen_data}. We provide visualizations from two examples here.

\textbf{Example 1:} Figure~\ref{fig:attn_ex1} shows a visualization of a sentence from the city of Asuit, Egypt, that the model correctly predicted. \textbf{Left:} Attention layer $\#3$ of the model\footnote{Layer counting starts from zero.} has several heads attending to lexical micro-dialectal cues related to Asuit. Most notably, tokens characteristic of the language of the correct city are attended to. Namely, the word
\<مطنك> (part of the metaphorical expression meaning ``what a devil") recieves attention in heads 1-3, and the word \<الرادل> ``man" in city of Asuit) receives attention in head 2. These cues usually co-occur with the token \<غور> (``you screwed [somebody]"), which is also characteristic of the Southern Egyptian region, and the city of Asuit. This is clear in the image in the \textbf{right} where the token \<غور> attends to other micro-dialectal cues in the sequence.

\textbf{Example 2:} 
Figure~\ref{fig:attn_ex2} shows a visualization of a sentence from the city of Marsa Matrouh, Egypt, that was incorrectly predicted as Tobruq, Libya. Even though the model makes a prediction error here, its error is meaningful in that it chooses a city that is located in the vicinity of that of the gold city. This means, interestingly, that the city-level model can pick a city close-by to gold in a different country rather than a far city in the same country. This reflects how micro-dialects paint a more nuanced (and linguistically plausible) picture. This also suggests that country-level dialect models are based on arbitrary assumptions, by virtue of being dependent on political boundaries which are not always what defines language variation. 

\subsection{Brief Error Analysis}\label{sec:append:error_analysis}
We provide a brief error analysis of single Task MARBERT (described in Table~\ref{tab:ourLMs} of the paper) in Table~\ref{tab:city_confusion}.

\begin{table*}[ht]
\centering
\resizebox{\textwidth}{!}{%
\begin{tabular}{llrl}
\hline
\multicolumn{1}{c}{\textbf{City}} & \multicolumn{1}{c}{\textbf{Country}} & \multicolumn{1}{c}{\textbf{Avg Error Dist}} & \multicolumn{1}{c}{\textbf{Countries with Confused Cities}}                                   \\\hline
Beni\_Malek              & Morocco                           & 4491.90                               & Oman 39.08, UAE 33.19, Morocco 18.07, Saudi Arabia 4.62, Libya 1.26   \\
Jabria                   & Morocco                           & 4331.06                              & UAE 22.63, Kuwait 20.44, Oman 10.95, Saudi Arabia 8.76, Libya 4.38    \\
Nouakchott               & Mauritania                        & 4324.06                              & Oman 14.36, Libya 13.81, Palestine 8.29, Algeria 8.29, Syria 7.18     \\
Casablanca               & Morocco                           & 3946.58                              & Oman 33.69, UAE 20.22, Morocco 7.68, Algeria 6.87, Libya 4.45         \\
Bordj\_El\_Kiffan        & Algeria                           & 3454.71                              & Kuwait 21.64, UAE 20.15, Bahrain 9.95, Oman 8.46, Saudi Arabia 6.97   \\
Ain\_Taya                & Algeria                           & 3318.77                              & Kuwait 17.24, Saudi Arabia 16.55, Oman 11.03, UAE 7.59, Tunisia 5.52  \\
Murzuq                   & Libya                             & 3255.02                              & Oman 40.0, UAE 28.57, Morocco 19.29, Libya 3.57, Syria 1.43          \\
Ouillen                  & Algeria                           & 3222.48                              & Saudi Arabia 20.55, UAE 13.7, Oman 12.33, Iraq 8.22, Yemen 8.22       \\
Laayoune                 & Morocco                           & 3146.83                              & Morocco 33.53, UAE 8.76, Algeria 8.16, Saudi Arabia 7.85, Oman 7.25  \\
Atar                     & Mauritania                        & 3109.41                              & Algeria 14.58, Morocco 14.58, Libya 10.42, Syria 10.42, Bahrain 6.25  \\
Ben\_Allel               & Algeria                           & 3027.69                              & Oman 26.0, Libya 14.0, Saudi Arabia 14.0, Algeria 10.0, Morocco 10.0  \\
Beni\_Mellal             & Morocco                           & 3019.40                               & Palestine 12.09, UAE 12.09, Morocco 10.99, Oman 9.89, Algeria 8.79    \\
Hargeysa                 & Somalia                           & 3011.22                              & Algeria 30.47, Saudi Arabia 10.16, UAE 10.16, Iraq 8.59, Morocco 7.81 \\
Dakhla                   & Morocco                           & 2924.22                              & Algeria 19.27, Morocco 15.27, Oman 9.09, Libya 8.36, Palestine 8.36  \\
Mogadishu                & Somalia                           & 2918.91                              & Saudi Arabia 18.98, Kuwait 11.92, Oman 9.98, Libya 7.54, Syria 7.54  \\
Bab\_Ezzouar             & Algeria                           & 2907.83                              & Iraq 46.67, Algeria 29.33, Somalia 10.67, Saudi Arabia 5.33          \\
Timimoun                 & Algeria                           & 2788.69                              & Algeria 22.89, Oman 13.25, Lebanon 7.23, Saudi Arabia 7.23, UAE 7.23  \\
Mohammedia               & Morocco                           & 2745.87                              & Morocco 16.81, Algeria 15.93, Oman 9.73, UAE 7.96, Libya 7.08         \\
Bni\_Oulid               & Morocco                           & 2702.39                              & Libya 32.35, Saudi Arabia 11.76, Morocco 8.82, Yemen 8.82, Oman 5.88  \\
El\_Jadida               & Morocco                           & 2693.21                              & Algeria 16.67, Morocco 11.11, Oman 11.11, Bahrain 11.11, Egypt 11.11 \\
\hline
\end{tabular}%
}
\caption{Top wrongly predicted cities in our DEV based on mBERT. For each gold city, we provide the average distance from the cities with which they were confused (we call it \textit{avg. error distance}), countries to which confused cities belong, followed by percentage in which cities of each country were confused with the gold city. In the future, we also plan to carry out a more extensive (including manual) error analysis based on the tweets involved.}
\label{tab:city_confusion}
\end{table*}

\begin{table}[]
\footnotesize
\begin{tabular}{@{}lrrccrrccrrcc@{}}
\toprule
                         & \multicolumn{4}{c}{\textbf{City}}                                                                         & \multicolumn{4}{c}{\textbf{State}}                                                                        & \multicolumn{4}{c}{\textbf{Country}}                                                                      \\ \midrule
                         & \multicolumn{1}{c}{\textbf{Accuracy}} & \multicolumn{1}{c}{\textbf{F1}} & \textbf{Epoch} & \textbf{Split} & \multicolumn{1}{c}{\textbf{Accuracy}} & \multicolumn{1}{c}{\textbf{F1}} & \textbf{Epoch} & \textbf{Split} & \multicolumn{1}{c}{\textbf{Accuracy}} & \multicolumn{1}{c}{\textbf{F1}} & \textbf{Epoch} & \textbf{Split} \\
\textbf{Narrow}          & 12.09                               & 8.78                          & 7              & A              & 16.11                               & 10.90                         & 4              & B              & 56.39                               & 41.37                         & 4              & B              \\
\textbf{Medium} & 7.04                                & 6.14                          & 7              & C              & 11.26                               & 7.48                          & 10             & B              & 46.34                               & 34.63                         & 4              & B              \\
\textbf{Wide}            & 5.27                                & 3.80                          & 9              & C              & 10.02                               & 4.96                          & 6              & A              & 42.87                               & 30.39                         & 3              & B              \\ \bottomrule
\end{tabular}
\end{table}